\documentclass[runningheads]{llncs}
\usepackage{graphicx}
\usepackage{comment}
\usepackage{amsmath,amssymb} 
\usepackage{color}

\usepackage{subfig}
\usepackage{float}
\usepackage{tabu,multirow}
\newcommand{\deflen}[2]{%
    \expandafter\newlength\csname #1\endcsname
    \expandafter\setlength\csname #1\endcsname{#2}%
}
\usepackage[table]{xcolor}

\usepackage{array}
\usepackage{chapterbib}

\begin{document}
\pagestyle{headings}
\mainmatter
\def\ECCVSubNumber{1358}  

\hbadness=99999
\hfuzz=999pt

\title{Adapting Object Detectors with Conditional Domain Normalization}

\titlerunning{Adapting Object Detectors with Conditional Domain Normalization}
%
\author{ Peng Su\inst{1,2} \and
Kun Wang\inst{2} \and
Xingyu Zeng\inst{2} \and
Shixiang Tang\inst{2} \\
Dapeng Chen\inst{2} \and
Di Qiu\inst{2} \and
Xiaogang Wang\inst{1}
}
\authorrunning{Peng Su et al.}
%
\institute{
The Chinese University of Hong Kong \and
SenseTime Research \\
\email{\{psu,xgwang\}@ee.cuhk.edu.hk} \\
}
\maketitle

\begin{abstract}
Real-world object detectors are often challenged by the domain gaps between different datasets.
In this work, we present the Conditional Domain Normalization (CDN) to bridge the domain gap.
CDN is designed to encode different domain inputs into a shared latent space, where the features from different domains carry the same domain attribute. To achieve this, we first disentangle the domain-specific attribute out of the semantic features from one domain via a domain embedding module, which learns a domain-vector to characterize the corresponding domain attribute information. Then this domain-vector is used to encode the features from another domain through a conditional normalization, resulting in different domains' features carrying the same domain attribute. We incorporate CDN into various convolution stages of an object detector to adaptively address the domain shifts of different level's representation. In contrast to existing adaptation works that conduct domain confusion learning on semantic features to remove domain-specific factors, CDN aligns different domain distributions by modulating the semantic features of one domain conditioned on the learned domain-vector of another domain. Extensive experiments show that CDN outperforms existing methods remarkably on both real-to-real and synthetic-to-real adaptation benchmarks, including 2D image detection and 3D point cloud detection.
\end{abstract}

\section{Introduction}
Deep neural networks have achieved remarkable success on visual recognition tasks~\cite{girshick2014rich,he2016deep}.
However, it is still very challenging for deep networks to generalize on a different domain, whose data distribution is not identical with original training data.
Such a problem is known as dataset bias or domain shift~\cite{quionero2009dataset}.
For example, to guarantee safety in autonomous driving, the perception model is required to perform well under all conditions, like sunny, night, rainy, snowy, etc.
However, even top-grade object detectors still face significant challenges when deployed in such varying real-world settings.
Although collecting and annotating more data from unseen domains can help, it is prohibitively expensive, laborious and time-consuming.
Another appealing application is to adapt from synthetic data to real data, as it can save the amount of cost and time.
However, current objector detectors trained with synthetic data can rarely generalize on real data due to a significant domain distribution gap~\cite{Peng2018Syn2RealAN,sankaranarayanan2018learning,tobin2017domain}.

Adversarial domain adaptation emerges as a hopeful method to learn transferable representations across domains.
It has achieved noticeable progress in various machine learning tasks, from image classification~\cite{liu2019compound,long2018conditional,Peng_2019_ICCV}, semantic segmentation ~\cite{sankaranarayanan2018learning,tsai2019domain,zou2018unsupervised}, object detection~\cite{Saito_2019_CVPR,Zhu_2019_CVPR} to reinforcement learning~\cite{james2019sim,peng2018sim,tobin2017domain}.
According to Ben-David's theory~\cite{ben2010theory}, the empirical risk on the target domain is bounded by the source domain risk and the $ \mathcal{H}$  domain divergence.
Adversarial adaptation dedicates to learn domain invariant representation to reduce the $ \mathcal{H}$  divergence, which eventually decreases the upper bound of the empirical error on the target domain.

However, existing adversarial adaptation methods still suffer from several problems.
First, previous methods~\cite{chen2018domain,ganin2015unsupervised,tobin2017domain} directly feed semantic features into a domain discriminator to conduct domain confusion learning.
But the semantic features contain both image contents and domain attribute information.
It's difficult to make the discriminator only focusing on removing domain-specific information without inducing undesirable influence on the images contents.
Second, existing adversarial adaptation methods~\cite{chen2018domain,ganin2015unsupervised,tobin2017domain}
use domain confusion learning at one or few convolution stages to handle the distribution mismatch, which ignores the differences of domain shifts at various representation levels.
For example, the first few convolution layers' features mainly convey low-level information of local patterns, while the higher convolution layers' features include more abstract global patterns with semantics~\cite{yosinski2015understanding}.
Such differences born within deep convolution neural networks naturally exhibit different types of domain shift at various convolution stages.

Motivated by this, we propose the Conditional Domain Normalization (CDN) to embed different domain inputs into a shared latent space, where the features of all different domains inputs carry the same domain attribute information.
Specifically,  CDN utilizes a domain embedding module to learn a domain-vector to characterize the domain attribute information, through disentangling the domain attribute out of the semantic features of domain inputs.
We use this domain-vector to encode the semantic features of another domain's inputs via a conditional normalization.
Thus different domain features carry the same domain attributes information.
We adopt CDN in various convolution stages to address different types of domain shift adaptively.
The experiment on both real-to-real and synthetic-to-real adaptation benchmarks demonstrate that our method outperforms the-state-of-the-art adaptation methods.
To summarize, our contributions are three folds:
(1) We propose the Conditional Domain Normalization to bridge the domain distribution gap, through embedding different domain inputs into a shared latent space, where the features from different domains carry the same domain attribute.
(2) CDN achieves state-of-the-art unsupervised domain adaptation performance on both real-to-real and synthetic-to-real benchmarks, including  2D image and 3D point cloud detection tasks.
And we conduct both quantitative and qualitative comparisons to analyze the features learned by CDN.
(3) We construct a large-scale synthetic-to-real driving benchmark for 2D object detection, including a variety of public datasets.

\section{Related work} \label{gen_inst}

\textbf{Object Detection} is the center topic in computer vision, which
is crucial for many real-world applications, such as autonomous driving.
In 2D detection, following the pioneering work of RCNN~\cite{girshick2014rich}, a number of object detection frameworks based
on convolutional networks have been developed like Fast R-CNN~\cite{girshick2015fast}, Faster R-CNN~\cite{ren2015faster}, and Mask R-CNN~\cite{he2017mask}, which significantly push forward the state of the art.
In 3D detection, spanning from detecting 3d objects from 2d images~\cite{chen20153d}, to directly generate 3D box from point cloud~\cite{qi2017pointnet++,shi2019pointrcnn}, abundant works has been successfully explored.
All these 2D and 3D objectors have achieved remarkable success on one or few specific public datasets.
However, even top-grade object detectors still face significant challenges when deployed in real-world settings.
The difficulties usually arise from the changes in environmental conditions.

\noindent \textbf{Domain Adaptation}
generalizes a model across different domains, and it has been extensively explored in various tasks, spanning from image  classification~\cite{bousmalis2017unsupervised,tzeng2017adversarial,long2018conditional,Peng_2019_ICCV,Liu_2020_CVPR}, semantic segmentation \cite{hoffman2017cycada,tsai2019domain,sankaranarayanan2018learning} to reinforcement learning~\cite{tobin2017domain,peng2018sim,james2019sim}.
For 2D detection, domain confusion learning via a domain discriminator has achieved noticeable progress in cross-domain detection.
\cite{chen2018domain} incorporated a gradient reversal layer~\cite{ganin2015unsupervised} into a Faster R-CNN model.
\cite{Saito_2019_CVPR,Zhu_2019_CVPR} adopt domain confusion learning on both global and local levels to align source and target distributions.
In contrast to existing methods conducting domain confusion learning directly on semantic features,
we explicitly disentangle the domain attribute out of semantic features.
And this domain attribute is used to encode other domains' features, thus different domain inputs share the same domain attribute in the feature space.
For 3D detection,
only a few works~\cite{yue2018lidar,hurl2019precise} has been explored to adapt object detectors across different point cloud dataset.
Different from existing works~\cite{yue2018lidar,hurl2019precise} are specifically designed for point cloud data, our proposed CDN is a general adaptation framework that adapts both 2D image and 3D point cloud object detector through the conditional domain normalization.

\noindent \textbf{Conditional Normalization}
is a technique to modulate the neural activation using a transformation that depends on external data.
It has been successfully used in the generative models and style transfer, like conditional batch normalization~\cite{dumoulin2016learned}, adaptive instance normalization (AdaIN)~\cite{huang2017arbitrary} and spatial adaptive batch normalization~\cite{park2019semantic}.
\cite{huang2017arbitrary} proposes AdaIN to control the global style of the synthesized image.
\cite{wang2018recovering} modulates the features conditioned on semantic masks for image  super-resolution.
\cite{park2019semantic} adopts a spatially-varying transformation, making it suitable for image synthesis from semantic masks.
Inspired by these works, we propose Conditional Domain Normalization (CDN) to modulate one domain's inputs condition on another domain's attributes information.
But our method exhibits significant difference with style transfer works:
Style transfer works modify a content image conditioned on another style image, which is a conditional instance normalization by nature;
but CDN  modulates one domain's features conditioned on the domain embedding learned from another domains' inputs (a group of images), which is like a domain-to-domain translation.
Hence we use different types of conditional normalization to achieve different goals.

\section{Method}
We first introduce the general unsupervised domain adaptation approach in section~\ref{sec: 3.1}.
Then we present the proposed Conditional Domain Normalization (CDN) in section~\ref{sec: 3.2}.
Last we adapt object detectors with the CDN in section~\ref{sec: 3.3}.

\subsection{General Adversarial Adaptation Framework} \label{sec: 3.1}
Given source images and labels $\{ (x_i^S, y_i^S) \}_{i=1}^{N_S}$ drawn from $P_{s}$, and target images $\{ x_i^T \} _{i=1}^{N_T}$ from target domain $P_{t}$,
the goal of unsupervised domain adaptation is to find a function $f: x \rightarrow y$ that minimize the empirical error on target data.
For object detection task, the $f$ can be decomposed as $f = G(\cdot; \theta_g) \circ H(\cdot; \theta_h) $, where $G(\cdot; \theta_g)$ represents a feature extractor network and $H(\cdot; \theta_h)$ denotes a bounding box head network.
The adversarial domain adaptation introduces a discriminator network $D(\cdot; \theta_d)$ that tries to determine the domain labels of feature maps generated by $G(\cdot; \theta_g)$.
\begin{eqnarray}
\begin{aligned} \label{eq:1}
 \min_{\theta_g, \theta_h} \enskip \mathcal{L}_{det} & = \mathcal{L}_{cls}(G(x; \theta_g), H(x; \theta_h)) +
  \mathcal{L}_{reg}(G(x; \theta_g), H(x; \theta_h)) \\
 \min_{\theta_d} \max_{\theta_g} \enskip  \mathcal{L}_{adv} &= \mathbb{E}_{x \sim P_{s}}[\log(D(G(x; \theta_g); \theta_d))] +
  \mathbb{E}_{x \sim P_{t}}[\log(1 - D(G(x; \theta_g); \theta_d)]
\end{aligned}
\end{eqnarray}
As illustrated in Eq.\ref{eq:1}, $G(\cdot; \theta_g)$ and $H(\cdot; \theta_h)$ are jointly trained to minimize the detection loss $\mathcal{L}_{det}$ by supervised training on the labeled source domain.
At the same time, the backbone $G(\cdot; \theta_g)$ is optimized to maximize the probability of $D(\cdot; \theta_d ) $ to make mistakes.
Through this two-player min-max game, the final $G(\cdot; \theta_g)$ will
converge to extract features that are indistinguishable for $D(\cdot; \theta_d ) $, thus domain invariant representations are learned.

\subsection{Conditional Domain Normalization} \label{sec: 3.2}

\begin{figure}[!t]
    \centering
    \includegraphics[width=0.98\textwidth]{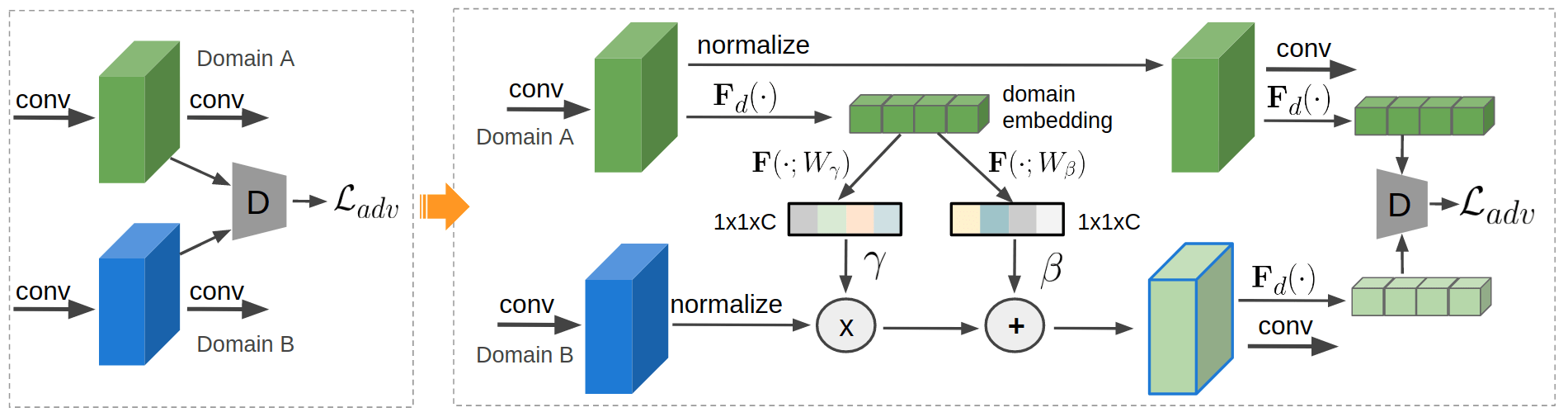}
    \caption{(Left) Traditional domain adversarial approach. (Right) Conditional Domain Normalization (CDN).
    The green and blue cubes represent the feature maps of domain A and domain B respectively.
    }
    \label{fig:CDN}
\end{figure}

Conditional Domain Normalization is designed to embed source and target domain inputs into a shared latent space, where
the semantic features from different domains carry the same domain attribute information.
Formally, let $v^s\in \mathbb{R}^{N \times C \times  H \times  W}$ and $v^t\in \mathbb{R}^ {N \times C \times  H \times  W}$ represent feature maps of source and target inputs, respectively.
$ C $ is the channel dimension and $N$ denotes the mini-batch size.
We first learn a domain embedding vector $e_{domain}^s \in \mathbb{R}^{1 \times C \times 1}$ to characterize the domain attribute of source inputs.
It is accomplished by a domain embedding network $\mathbf{F}_d (\cdot ; W)$ parameterized by two fully-connect layers with ReLU non-linearity $\delta$ as
\begin{equation}
    e_{domain}^s = \mathbf{F}_d( v_{avg}^{s}; W) = \delta (W_2 \delta ( W_1  v_{avg}^{s} )).
\end{equation}
And $v_{avg}^{s} \in \mathbb{R}^{N \times C \times 1} $ represents the channel-wise statistics of source feature $v^s$ generated by global average pooling
\begin{equation}
   v_{avg}^{s} =\frac{1}{HW} \sum_{h =1}^{H} \sum_{w =1}^{W}  v^s(h ,w).
\end{equation}

To embed both source and target domain inputs into a shared latent space, where source and target features carry the same domain attributes while preserving individual image contents.
We encode the target features $v^t$ with the source domain embedding via an affine transformation  as
\begin{equation}
    \hat v^t =  \mathbf{F}(e_{domain}^s; W_{\gamma}, b_\gamma) \cdot \left(  \dfrac{v^t - \mu^t}{\sigma^t} \right) +  \mathbf{F}(e_{domain}^s; W_{\beta}, b_\beta),
\end{equation}
where $\mu^t$ and $\sigma^t$ denote the mean and variance of target feature $v^t$.
The affine parameters are learned by function $F(\cdot ; W_{\gamma}, b_\gamma)$ and $ F(\cdot ; W_{\beta}, b_\beta)$ conditioned on the source domain embedding vector $e_{domain}^s$,
\begin{equation}
  F(e_{domain}^s; W_{\gamma}, b_\gamma) = W_{\gamma} e_{domain}^s + b_\gamma, \quad F(e_{domain}^s; W_{\beta},b_\beta) =  W_{\beta} e_{domain}^s + b_\beta.
\end{equation}

For the target feature mean  $\mu^t \in \mathbb{R}^{ 1 \times C  \times 1}  $ and variance  $\sigma^t \in \mathbb{R}^{ 1 \times C \times 1 } $, we calculate it with a standard batch normalization~\cite{ioffe2015batch}
\begin{equation}
    \mu_c^t = \frac{1}{NHW} \sum_{n =1}^{N} \sum_{h =1}^{H} \sum_{w =1}^{W}  v_{nchw}^t,  \quad
    \sigma_c^t = \sqrt {\frac{1}{NHW}  \sum_{n =1}^{N} \sum_{h =1}^{H} \sum_{w =1}^{W} (v_{nchw}^t - \mu_c^t)^2 + \epsilon},
\end{equation}
where $\mu_c^t$ and $\sigma_c^t$ denotes $c$-th channel of $ \mu^t$ and $\sigma^t$. Finally, we have a discriminator to supervise the encoding process of domain attribute as
\begin{eqnarray}
\begin{aligned} \label{eq:CDN_loss}
\min_{\theta_d} \max_{\theta_g} \enskip & \mathcal{L}_{adv} = \mathbb{E}[\log(D(\mathbf{F}_d(v^s)); \theta_d)] +
\mathbb{E}[\log(1 - D(\mathbf{F}_d(\hat v^t); \theta_d))],
\end{aligned}
\end{eqnarray}
where $v^s$ and $v^t$ are generated by $G(\cdot;\theta_g)$.

\textbf{Discussion}
CDN exhibits a significant difference compared with existing adversarial adaptation works.
As shown in Fig.~\ref{fig:CDN},
previous methods conduct domain confusion learning directly on semantic features to remove domain-specific factors.
However, the semantic features contain both domain attribute and image contents.
It is not easy to enforce the domain discriminator only regularizing the domain-specific factors without inducing any undesirable influence on image contents.
In contrast, we disentangle the domain attribute out of the semantic features via conditional domain normalization.
And this domain attribute is used to encode other domains' features, thus  different  domain features carry the same domain attribute information.

\begin{figure*}[!t]
\setlength{\textfloatsep}{8pt}
\begin{center}
  \includegraphics[width=0.98\textwidth]{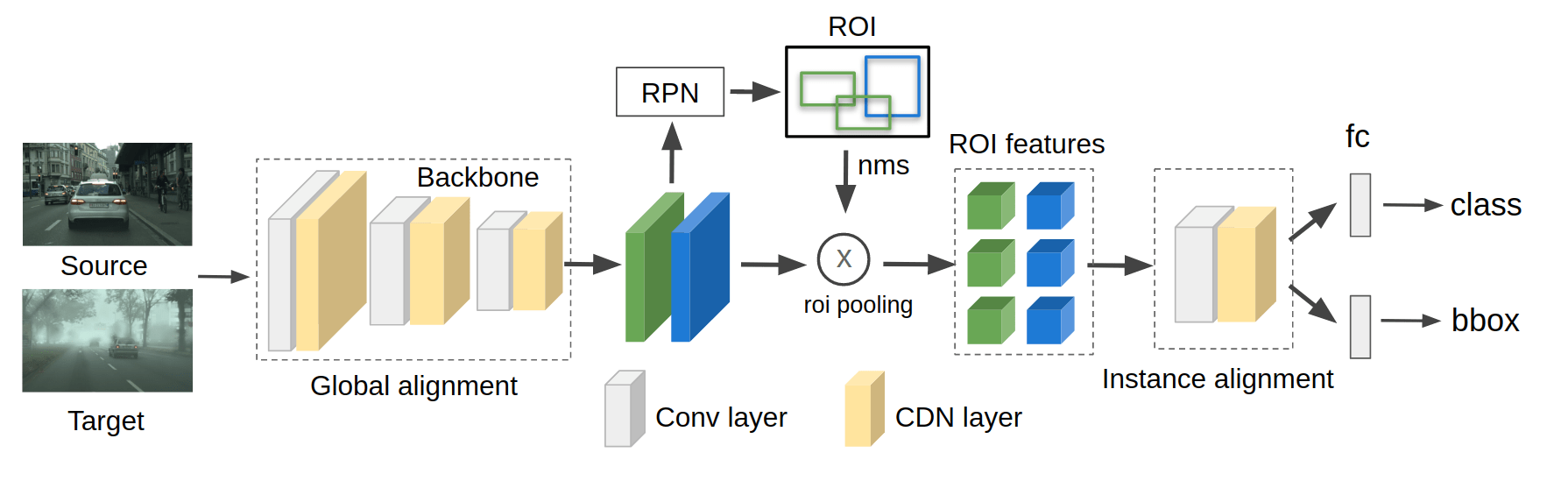}
  \caption{Faster R-CNN network incorporates with CDN.
  The CDN is adopted in both backbone network and bounding box head network to adaptively address the domain shift at different representation levels.
}
  \label{fig:Faster-CDN}
\end{center}
\end{figure*}

\subsection{Adapting Detector with Conditional Domain Normalization} \label{sec: 3.3}
Convolution neural network's (CNN) success in pattern recognition has been largely attributed to its great capability of learning hierarchical representations~\cite{yosinski2015understanding}.
More specifically, the first few layers of CNN focus on low-level features of local pattern, while higher layers capture semantic representations.
Given this observation, CNN based object detectors naturally exhibit different types of domain shift at various levels' representations.
Hence we incorporate CDN into different convolution stages in object detectors to address the  domain mismatch adaptively, as shown in Fig.\ref{fig:Faster-CDN}.

Coincident to our analysis, some recent works \cite{Saito_2019_CVPR,Zhu_2019_CVPR} empirically demonstrate that global and local region alignments have different influences on detection performance.
For easy comparison, we refer to the CDN located at the backbone network as global alignment, and CDN in the bounding box head networks as local or instance alignment.

As shown in Fig.~\ref{fig:Faster-CDN}, taking faster-RCNN model~\cite{ren2015faster} with ResNet~\cite{he2016deep} backbone as an example, we incorporate CDN in the last residual block at each stage.
Thus the global alignment loss can be computed by
\begin{equation}
    L_{adv}^{global} = \sum_{l =1}^{L} \mathbb{E}[\log(D_l(\mathbf{F}_d^l(v_l^s); \theta_d^l)] + \mathbb{E}[\log(1 - D_l(
    \mathbf{F}_d^l(\hat v^t_{l}) ; \theta_d^l))],
\end{equation}
where $v^s_{l}$ and $v^t_{l}$ denote $l$-th layer's source feature and the encoded target feature, and $D_l$ represents the corresponding domain discriminator parameterized by $\theta_d^l$.

As for bounding box head network, we adopt CDN on the fixed-size region of interest (ROI) features generated by ROI pooling~\cite{ren2015faster}.
Because the original ROIs are often noisy and the quantity of source and target ROIs are not equal,
we randomly select $\min (N_{roi}^S, N_{roi}^T)$ ROIs from each domain.
$N_{roi}^S$ and $N_{roi}^T$ represent the quantity of source and target ROIs after non-maximum suppression (NMS).
Hence we have instance alignment regularization for ROI features as
\begin{equation}
    L_{adv}^{instance} =  \mathbb{E}[\log(D_{roi}(\mathbf{F}_d^{roi}(v^s_{roi}); \theta_d^{roi})] + \mathbb{E}[\log(1 - D_{roi}(
    \mathbf{F}_d^{roi}(\hat v^t_{roi}); \theta_d^{roi}))].
\end{equation}

The overall training objective is to minimize the detection loss $\mathcal{L}_{det}$ (of the labeled source domain) that consists of a classification loss $\mathcal{L}_{cls}$ and a regression loss $\mathcal{L}_{reg}$, and min-max a adversarial loss $\mathcal{L}_{adv}$ of discriminator network
\begin{eqnarray}
\begin{aligned} \label{eq:overall_objective}
 \min_{\theta_d} \max_{\theta_g} \enskip  \mathcal{L}_{adv} &= \lambda  L_{adv}^{global}  + L_{adv}^{instance} \\
 \min_{\theta_g, \theta_h} \enskip \mathcal{L}_{det} & = \mathcal{L}_{cls}(G(x; \theta_g), H(x; \theta_h)) +
  \mathcal{L}_{reg}(G(x; \theta_g), H(x; \theta_h)),
\end{aligned}
\end{eqnarray}
where $\lambda$ is a weight to balance the global and local alignment regularization.


\section{Experiments}
We evaluate CDN on various real-to-real (KITTI to Cityscapes) and synthetic-to-real (Virtual KITTI/Synscapes/SIM10K to BDD100K, PreSIL to KITTI) adaptation benchmarks.
We also report results on cross-weather adaptation, Cityscapes to Foggy Cityscapes.
Mean average precision (mAP) with an intersection-over-union (IOU) threshold of $0.5$ is reported for 2D detection experiments.
We use Source and Target to represent the results of supervised training on source and target domain, respectively.
For 3D point cloud object detection,
PointRCNN~\cite{shi2019pointrcnn} with backbone of PointNet++~\cite{qi2017pointnet++} is adopted as our baseline model.
Following standard metric on KITTI benchmark~\cite{shi2019pointrcnn}, we use Average Precision(AP) with IOU threshold 0.7 for car and 0.5 for pedestrian/cyclist.

\subsection{Dataset}
{\bf Cityscapes}~\cite{cordts2016cityscapes}
is a European traffic scene dataset, which contains $2,975$ images for training and $500$ images for testing.

\noindent   {\bf Foggy Cityscapes}~\cite{SDV18} derives from Cityscapes with a fog simulation proposed by~\cite{SDV18}. It also includes $2,975$ images for training, $500$ images for testing.

\noindent  {\bf KITTI}~\cite{Geiger2012CVPR} contains $21,260$ images collected from different urban scenes, which includes 2D RGB images and 3D point cloud data.

\noindent  {\bf Virtual KITTI}~\cite{gaidon2016virtual} is derived from KITTI with a  real-to-virtual cloning technique proposed by \cite{gaidon2016virtual}. It has the same number of images and categories as KITTI.

\noindent  {\bf Synscapes}~\cite{wrenninge2018synscapes} is a synthetic dataset of street scene, which consists of $25,000$ images created with a photo-realistic rendering technique.

\noindent  {\bf SIM10K}~\cite{johnson2016driving} is a street view dataset generated from the realistic computer game Grand Theft Auto V (GTA-V).
It has $10,000$ training images and the same categories as in Cityscapes.

\noindent  {\bf PreSIL}~\cite{hurl2019precise} is synthetic point cloud dataset derived from GTA-V, which
consists of $50,000$ frames of high-definition images and point clouds.

\noindent  {\bf BDD100K} ~\cite{yu2018bdd100k} is a large-scale dataset (contains 100k images) that covers diverse driving scenes.
It is a good representative of real data in the wild.

\subsection{Implementation Details} \label{Implementation}
We train the Faster R-CNN~\cite{ren2015faster} model for 12 epochs on all experiments. The model is optimized by SGD with multi-step learning rate decay.
SGD uses the learning rate of 0.00625 multiplied by the batchsize, and momentum of 0.9.
We adopt CDN layer in all convolution stages, including the backbone and bounding box network.
All experiments use sync BN~\cite{peng2018megdet} with a batchsize of $32$.
$\lambda$ is set as 0.4 by default in all experiments.
On Cityscapes to Foggy Cityscapes adaptation, we follow~\cite{Saito_2019_CVPR,Zhu_2019_CVPR}
to prepare the train/test split, and use an image shorter side of 512 pixels.
On synthetic-to-real adaptation,
for a fair comparison, we randomly select $7000$ images for training and $3000$ for testing, for all synthetic datasets and BDD100K dataset.
For 3D point cloud detection, we use PointRCNN~\cite{shi2019pointrcnn} model with same setting as~\cite{shi2019pointrcnn}.
We incorporated the CDN layer in the point-wise feature generation stage (global alignment) and 3D ROIs proposal stage (instance alignment).

\section{Experimental Results and Analysis}

\subsection{Results on Cityscapes to Foggy Cityscapes} \label{sec:5-1}
We compare CDN with the state-of-the-art methods in Table~\ref{tb:cityscapes_to_foggy_cityscapes}.
Following \cite{Saito_2019_CVPR,Zhu_2019_CVPR}, we also report results using Faster R-CNN model with VGG16 backbone.
As shown in Table~\ref{tb:cityscapes_to_foggy_cityscapes},
CDN outperforms previous state-of-the-art methods by a large margin of $1.8\%$ mAP.
The results demonstrate the effectiveness of CDN on reducing domain gaps.
A detailed comparison of different CDN settings can be found at the ablation study~\ref{Ablation}.
As shown in Fig.~\ref{fig:syn2real}, our method exhibits good generalization capability under foggy weather conditions.

\begin{table*}[]
\centering
\begin{tabular}{l|cccccccc|c}
\hline
Method                              & Person & Rider & Car  & Truck  & Bus  & Train  & Motorcycle & Bicycle  &  mAP \\
\hline
Source                          & 29.3   & 31.9  &43.5  & 15.8   & 27.4  & 9.0   &20.3        & 29.9     & 26.1 \\
\hline
DA-Faster  \cite{chen2018domain}    & 25.0 & 31.0 & 40.5 & 22.1 & 35.3 & 20.2 & 20.0 & 27.1 & 27.9 \\
DT  \cite{Inoue_2018_CVPR}         & 25.4 & 39.3 & 42.4 & 24.9 & 40.4 & 23.1 & 25.9 & 30.4 & 31.5 \\
SCDA \cite{Zhu_2019_CVPR}         & 33.5 & 38.0 & 48.5 & 26.5 & 39.0 & 23.3 & 28.0 & 33.6 & 33.8 \\
DDMRL \cite{kim2019diversify}        & 30.8 & 40.5 & 44.3 & 27.2 & 38.4 & \bf 34.5 & 28.4 & 32.2 & 34.6 \\
SWDA \cite{Saito_2019_CVPR}       & 30.3 & 42.5 & 44.6 & 24.5 & 36.7 & 31.6 & 30.2 & 35.8 & 34.8 \\
\hline
CDN (ours)                      & \bf 35.8 &\bf 45.7 & \bf 50.9  & \bf 30.1  & \bf 42.5 & 29.8 & \bf 30.8 & \bf 36.5 & \bf 36.6 \\
\hline

\end{tabular}
\vspace{4pt}
\caption{Cityscapes to Foggy Cityscapes adaptation.
}
\label{tb:cityscapes_to_foggy_cityscapes}
\end{table*}


\subsection{Results on KITTI to Cityscapes}
Different camera settings may influence the detector performance in real-world applications.
We conduct the cross-camera adaptation on KITTI to Cityscapes.
Table~\ref{tab:kitti2city} shows the adaptation results on \emph{car} category produced by Faster R-CNN with VGG16.
Global and Instance represent global and local alignment respectively.
The results demonstrate that CDN achieves $1.7\%$ mAP improvements over the state-of-the-art methods.
We can also find that instance feature alignment contributes to a larger performance boost than global counterpart, which is consistent with previous discovery~\cite{Saito_2019_CVPR,Zhu_2019_CVPR}.


\subsection{Results on SIM10K to Cityscapes}
Following the setting of~\cite{Saito_2019_CVPR}, we evaluate the detection performance on \emph{car} on SIM10K-to-Cityscapes benchmark.
The results in Table~\ref{tab:sim2city} demonstrate CDN constantly performs better than the baseline methods.
CDN with both global and instance alignment achieves $49.3\%$ mAP on validation set of Cityscapes, which outperforms the previous state-of-the-art method by $1.6\%$ mAP.

\subsection{Results on Synthetic to Real Data} \label{sec:5-4}
\begin{figure*}[]
\centering
\subfloat{\includegraphics[width=2.3\widthdef]{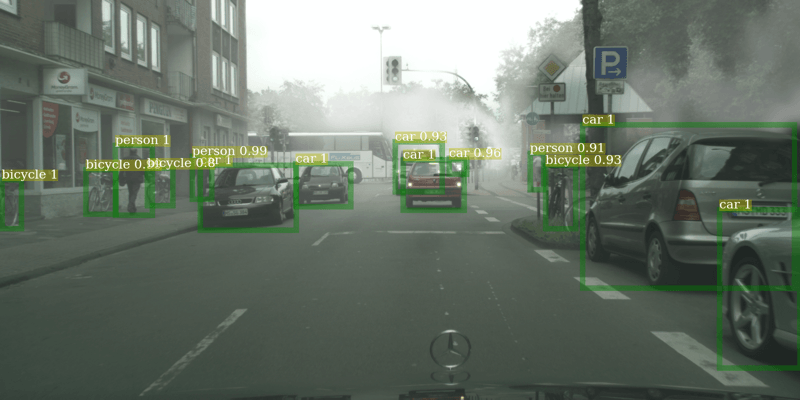}}  \hspace{0.5pt}
\subfloat{\includegraphics[width=2.3\widthdef]{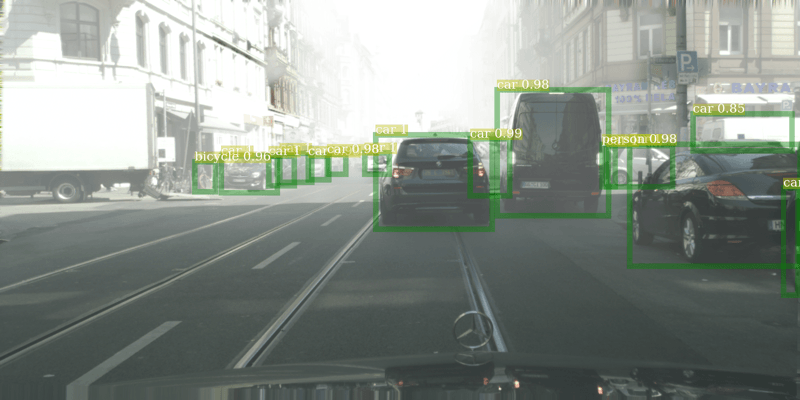}}     \hspace{0.5pt}
\subfloat{\includegraphics[width=2.3\widthdef]{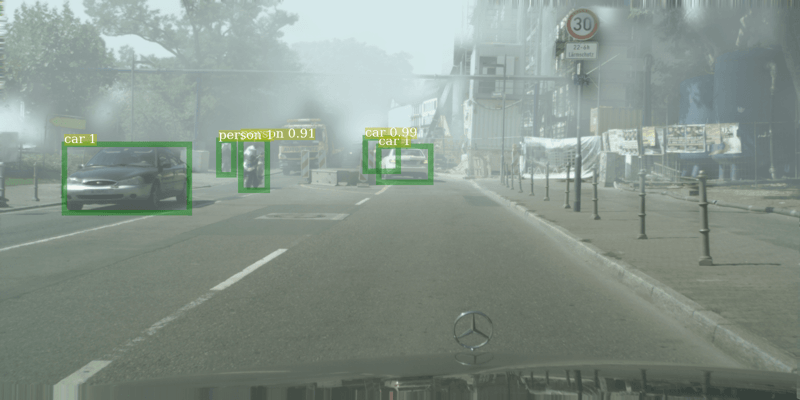}}    \hspace{0.5pt}
\subfloat{\includegraphics[width=2.3\widthdef]{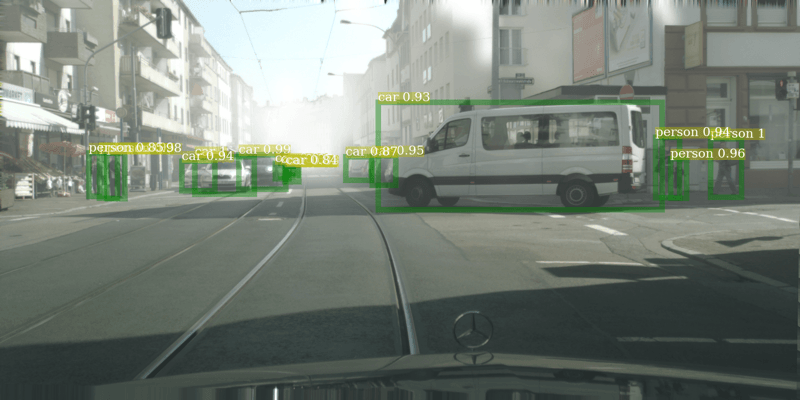}} \\

\subfloat{\includegraphics[width=2.3\widthdef]{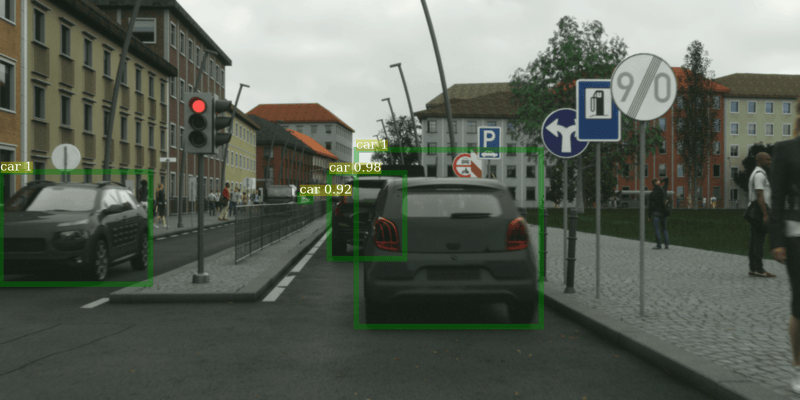}}
\hspace{0.5pt}
\subfloat{\includegraphics[width=2.3\widthdef]{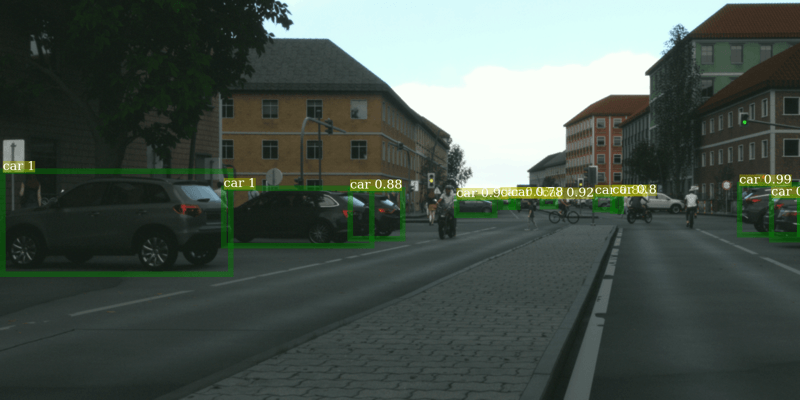}}     \hspace{0.5pt}
\subfloat{\includegraphics[width=2.3\widthdef]{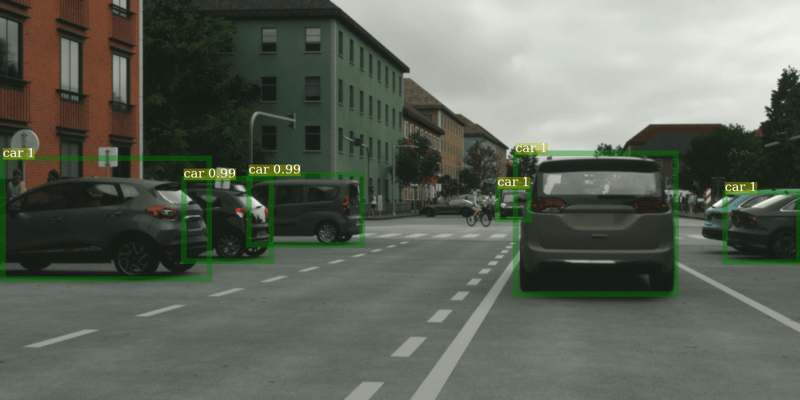}}    \hspace{0.5pt}
\subfloat{\includegraphics[width=2.3\widthdef]{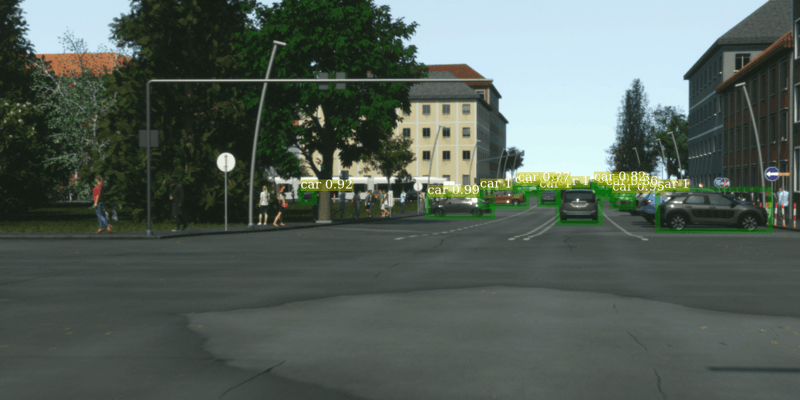}}
\\

\subfloat{\includegraphics[width=2.3\widthdef]{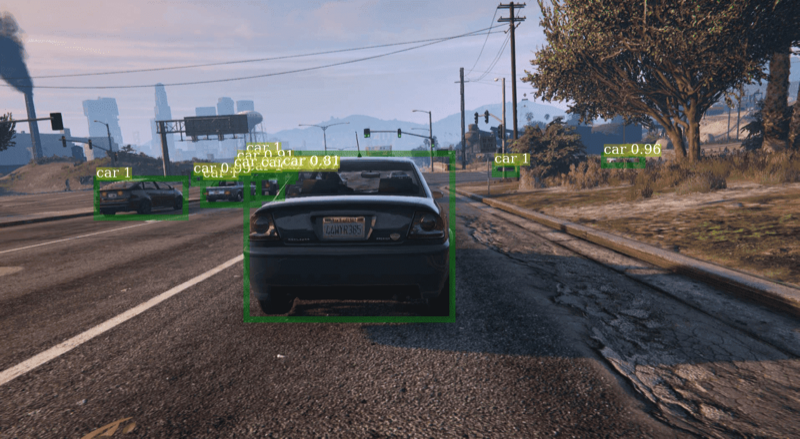}}  \hspace{1pt}
\subfloat{\includegraphics[width=2.3\widthdef]{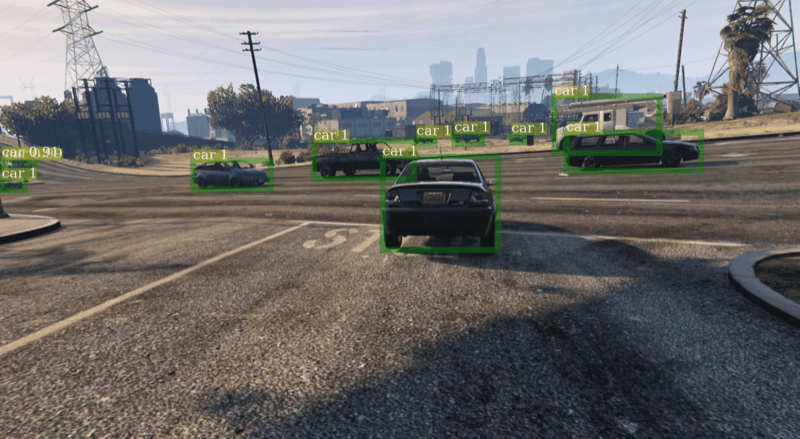}}     \hspace{1pt}
\subfloat{\includegraphics[width=2.3\widthdef]{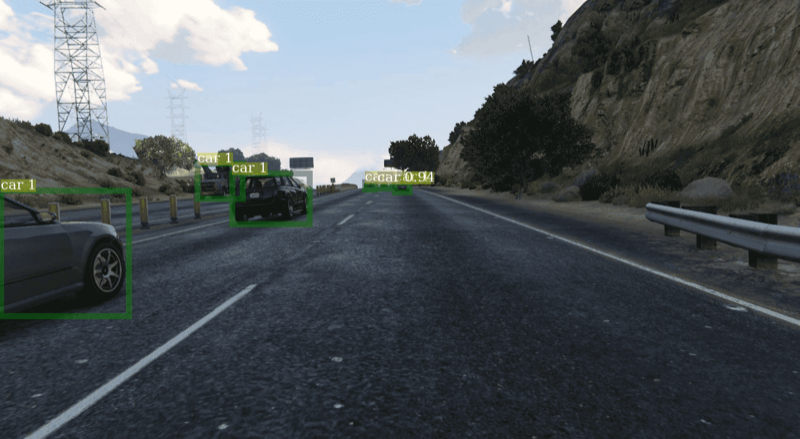}}    \hspace{1pt}
\subfloat{\includegraphics[width=2.3\widthdef]{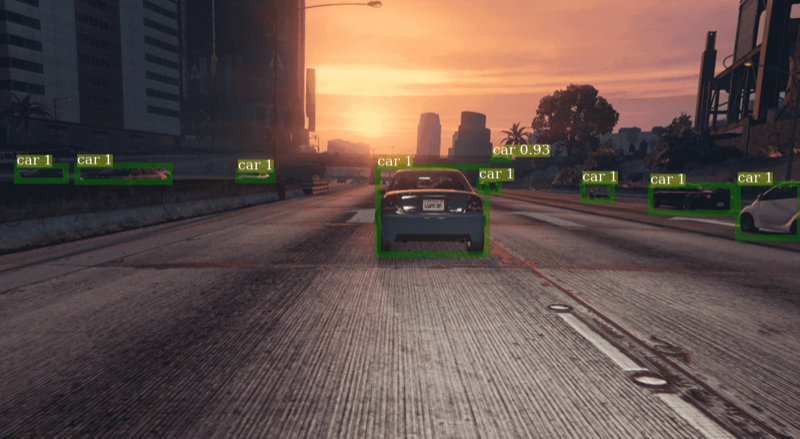}}    \hspace{1pt} \\

\subfloat{\includegraphics[width=2.3\widthdef]{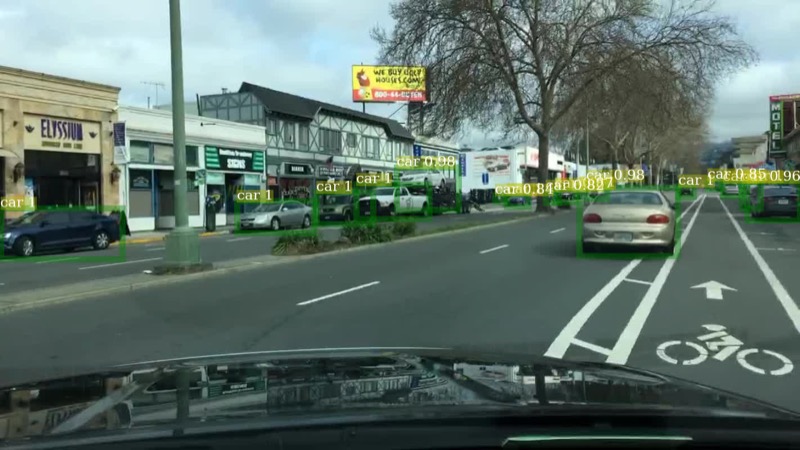}}    \hspace{1pt}
\subfloat{\includegraphics[width=2.3\widthdef]{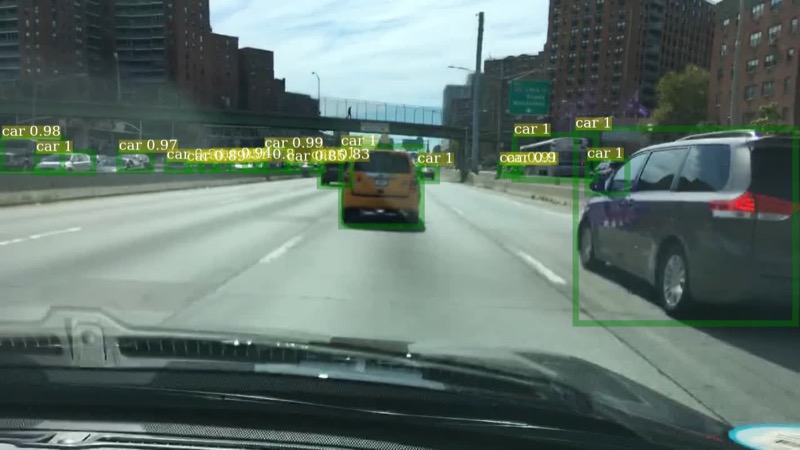}}    \hspace{1pt}
\subfloat{\includegraphics[width=2.3\widthdef]{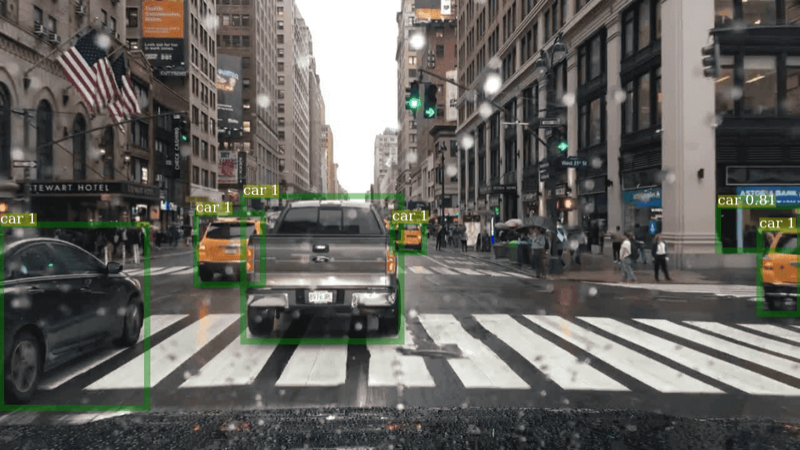}}    \hspace{1pt}
\subfloat{\includegraphics[width=2.3\widthdef]{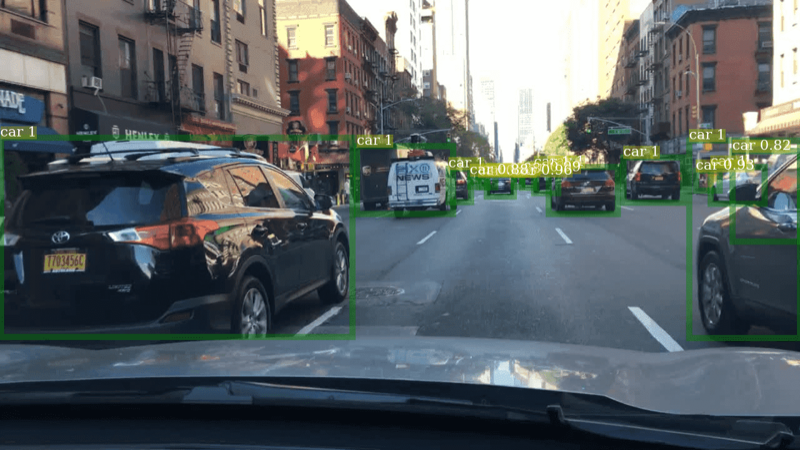}}    \hspace{1pt} \\
\caption{Example results on Foggy Cityscapes/Synscapes/SIM10K/BDD100K (from top to bottom). The results are produced by a Faster R-CNN model incorporated with CDN. The class and score predictions are at the top left corner of the bounding box. Zoom in to visualize the details.
 }
\label{fig:syn2real}
\end{figure*}

To thoroughly evaluate the performance of the state-of-the-art methods on synthetic to real adaptation,
we construct a large-scale synthetic-to-real adaptation benchmark on various public synthetic datasets, including Virtual KITTI, Synscapes and SIM10K.
``All''  represents using the combination of 3 synthetic datasets.
Compared with SIM10K-to-Cityscapes, the proposed benchmark is more challenging in terms of much larger image diversity in both real and synthetic domains.
We compare CDN with the state-of-the-art method SWDA\cite{Saito_2019_CVPR}  in Table~\ref{tb: syn2real}.
CDN consistently outperforms SWDA under different backbones, which achieves average $2.2\%$ mAP and $2.1\%$ mAP improvements on Faster-R18 and Faster-R50 respectively.
Using the same adaptation method, the detection performance strongly depends on the quality of synthetic data.
For instance, the adaptation performance of SIM10K is much better than Virtual KITTI.
Some example predictions produced by our method are visualized in Fig.~\ref{fig:syn2real}.

\begin{table}[]
\parbox{.5\linewidth}{
\centering
\begin{tabular}{|l|cc|c|}

        \hline
         Method    & Global \quad & Instance   & mAP(\%) \\
        \hline
         Source only    &              &        &   37.1     \\
         DA-Faster \cite{chen2018domain}     &      \checkmark           &   \checkmark       &  38.3       \\
         SWDA \cite{Saito_2019_CVPR}             &    \checkmark            &  \checkmark     &  43.2        \\
        SCDA \cite{Zhu_2019_CVPR}              &      \checkmark          &   \checkmark     &    42.9        \\
         \hline
         \multirow{3}{*}{CDN}           &     \checkmark           &      &   40.2        \\
                    &                &   \checkmark   &    43.1       \\
                     &       \checkmark         &  \checkmark    &  \bf   44.9      \\
        \hline
        \end{tabular}
\caption{KITTI to Cityscapes.}
\label{tab:kitti2city}
}
\hfill
\parbox{.48\linewidth}{
\centering
\begin{tabular}{|l|cc|c|}

        \hline
         Method    & Global \quad & Instance   & mAP(\%)   \\
        \hline
         Source only    &              &        &   34.3     \\
         DA-Faster\cite{chen2018domain}     &         \checkmark        &   \checkmark      &  38.3       \\
         SWDA \cite{Saito_2019_CVPR}             &      \checkmark          &    \checkmark    &    47.7        \\
        SCDA \cite{Zhu_2019_CVPR}              &     \checkmark           &   \checkmark     &    44.1        \\
         \hline
         \multirow{3}{*}{CDN}           &     \checkmark           &      &   41.2        \\
                    &                &   \checkmark   &    45.8       \\
                     &       \checkmark         &  \checkmark    &    \bf 49.3      \\
        \hline
        \end{tabular}
        \caption{SIM10K to Cityscapes.}
\label{tab:sim2city}
}
\end{table}

\vspace{-20pt}

\begin{table}[]
\centering
\begin{tabular}{l|l|c|c|c|c}
\hline
Model                       &  Method               & Virtual KITTI      &  Synscapes      &  SIM10K    & All  \\ \hline
\multirow{4}{*}{Faster-R18} & Source                &    9.8             &  24.5           &  37.7      &  38.2 \\
                      & SWDA\cite{Saito_2019_CVPR}  & 15.6               & 27.0            &  40.2      &   41.3    \\
                            & CDN                   &  \bf  17.5         & \bf  29.1    & \bf  42.7     &  \bf  43.6   \\
                            & Target          & \multicolumn{4}{c}{\cellcolor{gray!10} 70.5}   \\  \hline
\multirow{4}{*}{Faster-R50}  & Source             &   13.9              & 29.1             & 41.6       &  42.8 \\
                & SWDA\cite{Saito_2019_CVPR}     &   19.7               & 31.5             & 42.9       &  44.3    \\
                             & CDN              & \bf 21.8              & \bf 33.4    & \bf 45.3        & \bf 47.2    \\
                            & Target               &  \multicolumn{4}{c}{\cellcolor{gray!10}75.6}   \\   \hline

\end{tabular}
\caption{Adaptation from different synthetic data to real data.
mAP on car is reported on BDD100K validation.
The results of supervised training on BDD100K are highlighted in gray.
}
\label{tb: syn2real}
\end{table}

\deflen{widthdef}{35pt}

\vspace{-10pt}

\subsection{Adaptation on 3D Point Cloud  Detection}

We evaluate CDN on adapting 3D object detector from synthetic point cloud (PreSIL) to real point cloud data (KITTI).
PointRCNN~\cite{shi2019pointrcnn} with backbone of PointNet++~\cite{qi2017pointnet++} is adopted as our baseline model.
Following standard metric on KITTI benchmark~\cite{shi2019pointrcnn}, we use Average Precision(AP) with IOU threshold 0.7 for car and 0.5 for pedestrian / cyclist.
Table~\ref{tab:3d_det} shows that CDN constantly outperforms the state-of-the-art method PointDAN~\cite{qin2019pointdan} across all categories, with an average improvement of $1.9\%$ AP.
We notice that instance alignment contributes to a larger performance boost than global alignment.
It can be attributed by the fact that point cloud data spread over a huge 3D space but most information is stored in the local foreground points (see Fig.~\ref{fig:point-cloud-sample1}).
\vspace{8pt}

\begin{minipage}{\textwidth}
  \hspace{-0.5cm}  \begin{minipage}[]{0.3\textwidth}
    \includegraphics[width=0.8\linewidth]{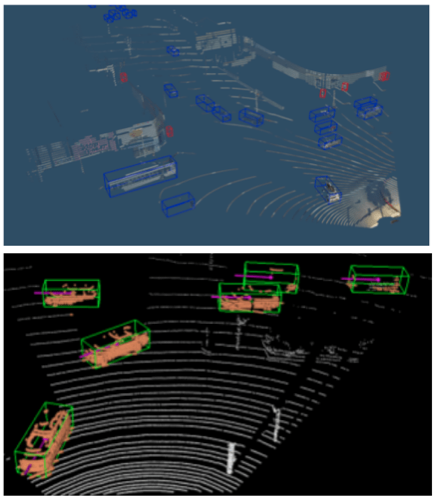}
    \captionof{figure}{Top:PreSIL; \\\hspace{\textwidth} Bottom:KITTI.}
    \label{fig:point-cloud-sample1}
  \end{minipage}
  \hspace{-0.4cm}
  \begin{minipage}[]{0.7\textwidth}
    \resizebox{0.95\textwidth}{!}{
    \begin{tabular}{|c|cc|c|c|c|}

        \hline
        Model                & Global      & Instance           &     Car         & Pedestrian      & Cyclist              \\
        \hline
        Source              &          &                        &    15.7           &   9.6              & 5.6     \\  
  CycleGAN\cite{saleh2019domain}   & \checkmark       &  \checkmark              &   16.5       &  10.3        & 5.9    \\
PointDAN\cite{qin2019pointdan}      &   \checkmark      &   \checkmark      &  17.1       &  10.9      & 7.5    \\
         \hline
 \multirow{3}{*}{ CDN}        &    \checkmark      &             &  17.3        &  10.5      &   6.0 \\
                           &          &   \checkmark             &  18.5        & 12.8       &  8.7  \\
                          &  \checkmark  & \checkmark          &  \bf  19.0      & \bf 13.2      & \bf 9.1    \\
        \hline
          Target           &          &        &   75.7    & 41.7      &  59.6   \\
        \hline
        \end{tabular}
    }
    \captionof{table}{Adapting from synthetic (PreSIL) to real (KITTI) pint cloud. AP of moderate level on KITTI test is reported.}
      \label{tab:3d_det}
    \end{minipage}

\end{minipage}

\section{Analysis}

\subsection{Visualize and Analyze the Feature Maps}
\begin{figure*}[]
\centering
\subfloat{\includegraphics[width=1.5\widthdef]{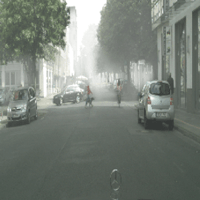}}  \hspace{0.15pt}
\subfloat{\includegraphics[width=1.5\widthdef]{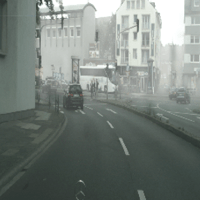}}   \hspace{0.15pt}
\subfloat{\includegraphics[width=1.5\widthdef]{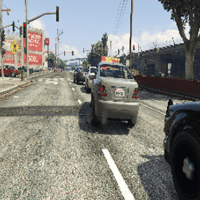}}  \hspace{0.15pt}
\subfloat{\includegraphics[width=1.5\widthdef]{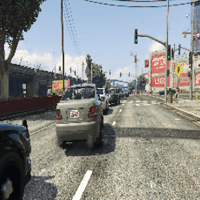}}   \hspace{0.15pt}
\subfloat{\includegraphics[width=1.5\widthdef]{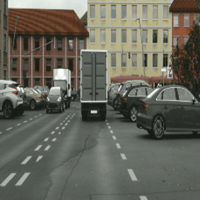}}  \hspace{0.15pt}
\subfloat{\includegraphics[width=1.5\widthdef]{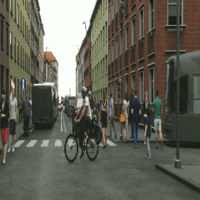}}
\\
\subfloat{\includegraphics[width=1.5\widthdef]{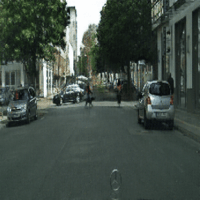}}  \hspace{0.15pt}
\subfloat{\includegraphics[width=1.5\widthdef]{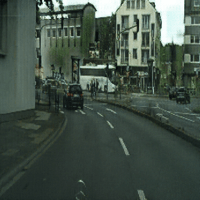}}   \hspace{0.15pt}
\subfloat{\includegraphics[width=1.5\widthdef]{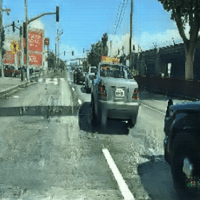}}  \hspace{0.15pt}
\subfloat{\includegraphics[width=1.5\widthdef]{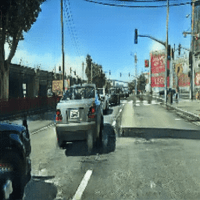}}   \hspace{0.15pt}
\subfloat{\includegraphics[width=1.5\widthdef]{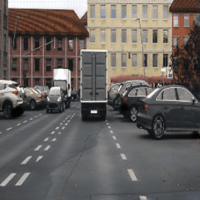}}  \hspace{0.15pt}
\subfloat{\includegraphics[width=1.5\widthdef]{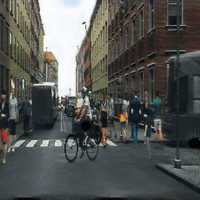}}
\\

\caption{Top row: Original inputs from Foggy Cityscapes, SIM10K  and Synscapes (left to right);
Bottom row: Reconstructed images from features encoded with the learned domain embedding of another domain.
}
\label{fig:domain_embedding}
\end{figure*}

Despite the general efficiency on various benchmarks, we are also interested in the underlying principle of CDN.
We interpret the learned domain embedding via appending a decoder network after the backbone to reconstruct the RGB images from the feature maps.
As shown in Fig.~\ref{fig:domain_embedding}, the top row shows the original inputs from Foggy Cityscapes, SIM10K  and Synscapes (left to right), and the bottom row shows the  reconstructed images from the corresponding features encoded with the domain embedding of another domain.
The reconstructed images carry the same domain style of another domain, suggesting the learned domain embedding captures the domain attribute information
and CDN can effectively transform the domain style of different domains.

Furthermore, we compute Fréchet Inception Distance (FID)\cite{heusel2017gans} to quantitatively investigate the difference between source and target features.
FID has been a popular metric to evaluate the style similarity between two groups of images in GANs.
Lower FID score indicates a smaller style difference.
For easy comparison, we normalize the FID score to $[0,1]$ by dividing the maximum score.
As shown in Table~\ref{tab:fid}, the feature learned with CDN achieves significantly smaller FID score compared with feature learned on source domain only, suggesting CDN effectively reduces the domain gap in the feature space.
Obviously, supervised joint training on source and target data gets the smallest FID score, which is verified by the best detection performance achieved by joint training.
As shown in Fig.~\ref{fig:fid_score}, synthetic-to-real has larger FID score than real-to-real dataset, since the former owns larger domain gaps.
\vspace{3pt}

\begin{minipage}{\textwidth}
  \hspace{-0.5cm} \begin{minipage}[]{0.48\textwidth}
    \centering
    \resizebox{\textwidth}{!}{
    \begin{tabular}{|c|*{4}{>{\centering\arraybackslash}p{1cm}|}}
    \hline
    \multirow{2}{*}{Method} & \multicolumn{2}{c|}{SIM to BDD}     & \multicolumn{2}{c|}{City to Foggy}   \\
    \cline{2-5}
                    &  FID    & mAP     & FID     & mAP          \\
    \hline
    Source          &  0.94   & 37.7    & 0.83     & 26.1    \\
    Joint training  & 0.67    & 79.3    & 0.41     &  49.5  \\
    SWDA~\cite{Saito_2019_CVPR}     & 0.83    & 40.2   & 0.76     & 34.8   \\
    CDN             & 0.71    & 42.7   & 0.60     & 36.6   \\
    \hline
    \end{tabular}
    }
    \captionof{table}{FID score and mAP.}
    \label{tab:fid}
    \end{minipage}
    \hfill
    \hspace{-0.55cm} \begin{minipage}[]{0.52\textwidth}
    \centering
    \includegraphics[width=0.85\linewidth]{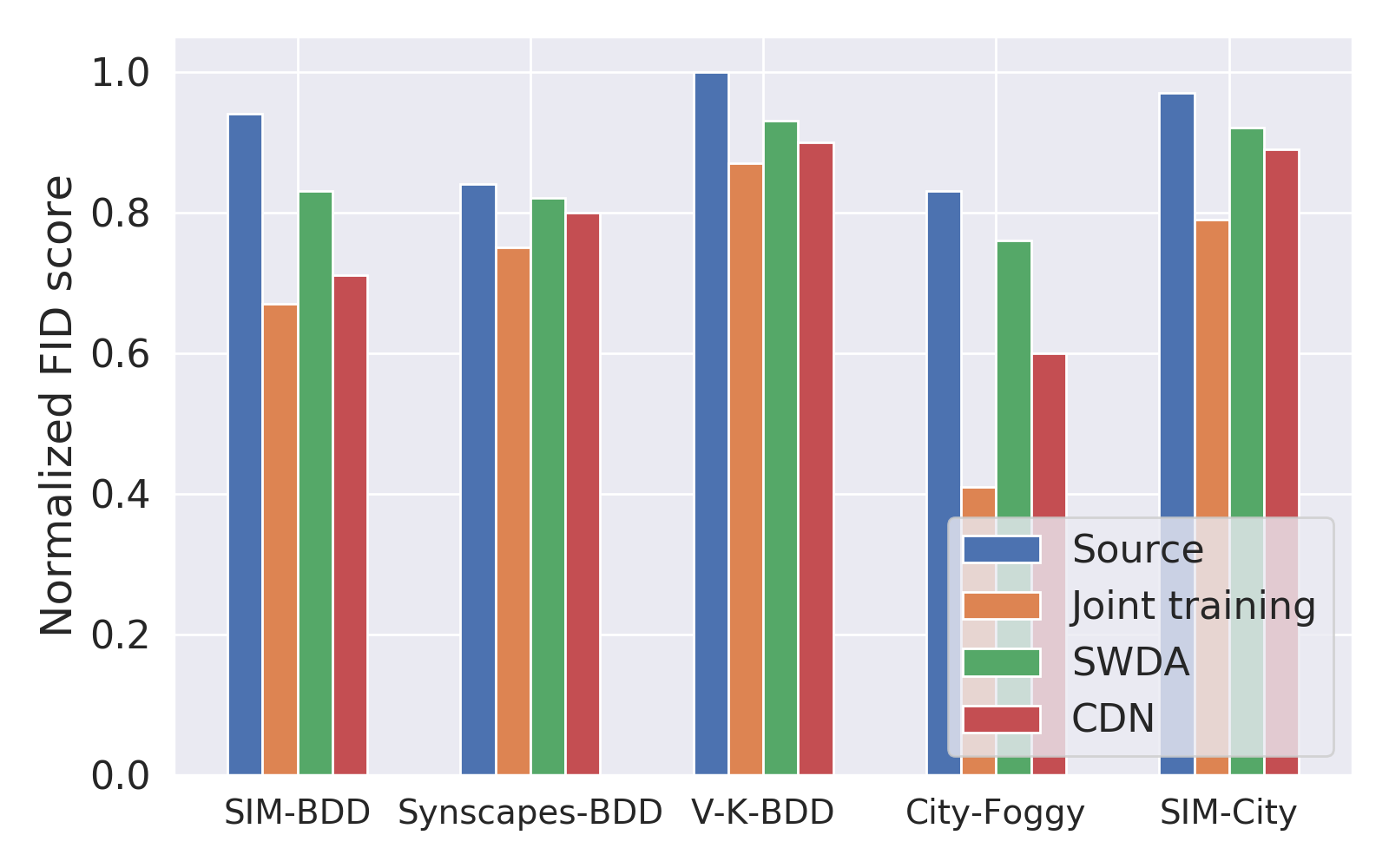}
    \vspace{-10pt}
    \captionof{figure}{FID scores on all datasets.}
    \label{fig:fid_score}
  \end{minipage}
\end{minipage}

\subsection{Analysis on Domain Discrepancy}
\begin{figure}[!th]
  \centering
    \subfloat[City-to-Foggy]{\includegraphics[width=0.25\linewidth]{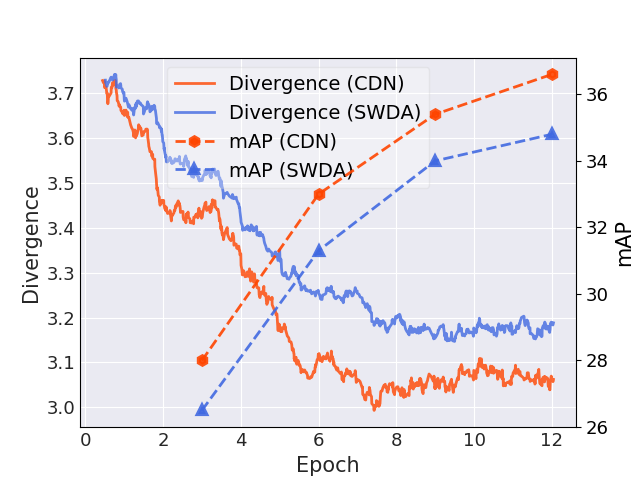}}
    \subfloat[City-to-Foggy]{\includegraphics[width=0.27\linewidth]{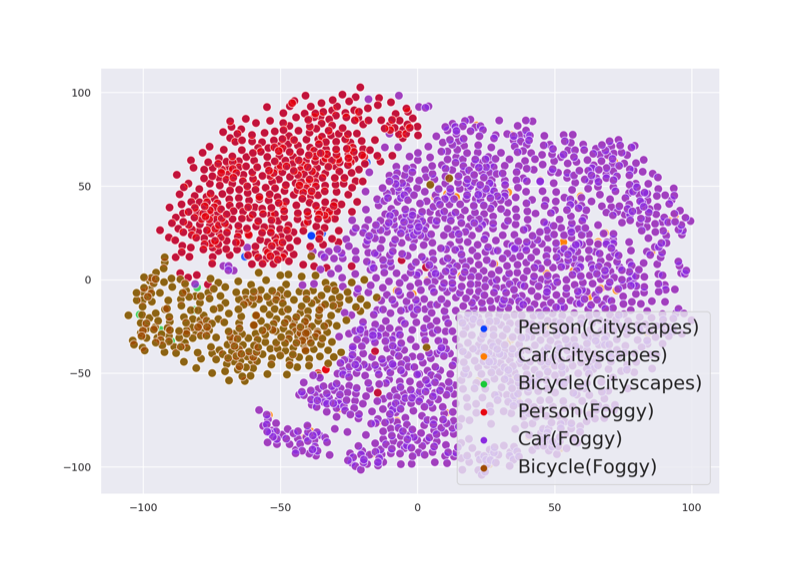}}
    \subfloat[SIM-to-BDD]{\includegraphics[width=0.25\linewidth]{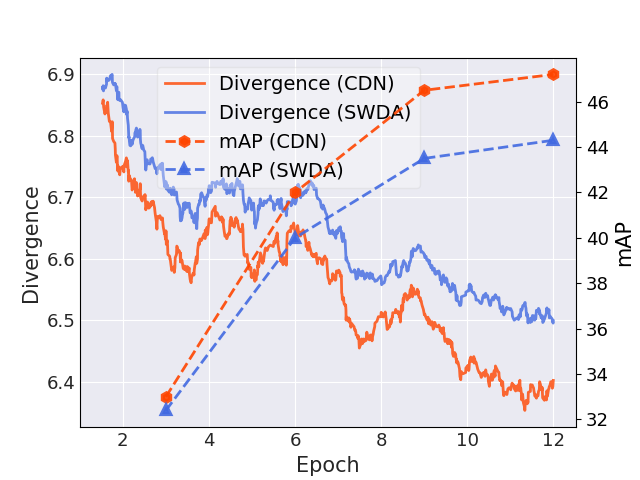}}
    \subfloat[SIM-to-BDD]{\includegraphics[width=0.27\linewidth]{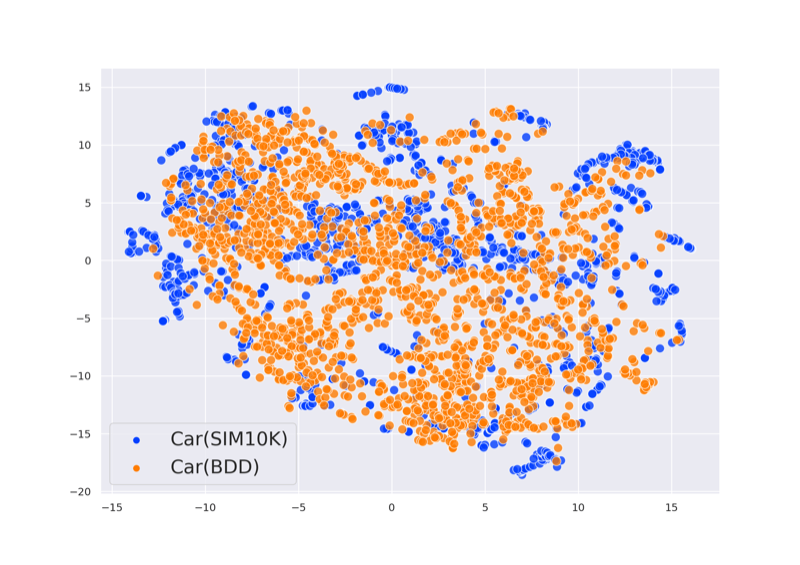}}
    \\ [-1.2ex]
  \caption{(a)(c): Divergence and adaptation performance. (b)(d): t-SNE plot of instance features.
  }
  \label{fig:vis_feat_divergence}
\end{figure}

We adopt symmetric Kullback–Leibler divergence to investigate the discrepancy between source and target domain in feature space.
To simplify the analysis, we assume source and target features are drawn from the multivariate normal distribution.
The divergence is calculated with the Res5-3 features and plotted in log scale.
Fig.~\ref{fig:vis_feat_divergence} (a) and (c) show that the domain divergence continues decreasing during training, indicating the Conditional Domain Normalization keeps reducing domain shift in feature space.
Benefiting from the reduction of domain divergence, the adaptation performance on the target domain keeps increasing.
Comparing with SWDA, CDN achieves lower domain discrepancy and higher adaptation performance.

Except for the quantitative measure of domain divergence, we also visualize the t-SNE plot of instance features extracted by a Faster R-CNN incorporated with CDN.
Fig.~\ref{fig:vis_feat_divergence} (b)(d) shows the t-SNE plot of instance features extracted by a Faster R-CNN model incorporated with CDN.
The same category features from two domains group in tight clusters, suggesting source and target domain distributions are well aligned in feature space.
Besides, features of different categories own clear decision boundaries, indicating discriminative features are learned by our method. These two factors contribute to the detection performance on target domain.

\section{Ablation Study}\label{Ablation}
For the ablation study, we use a Faster R-CNN model with ResNet-18 on SIM10K to BDD100K adaptation benchmark, and a Faster R-CNN model with VGG16 on Cityscapes-to-Foggy Cityscapes adaptation benchmark.
G and I denote adopting CDN in the backbone and bounding box head network, respectively.
\begin{figure}[]
  \centering
    \subfloat[]{\includegraphics[width=0.38\linewidth]{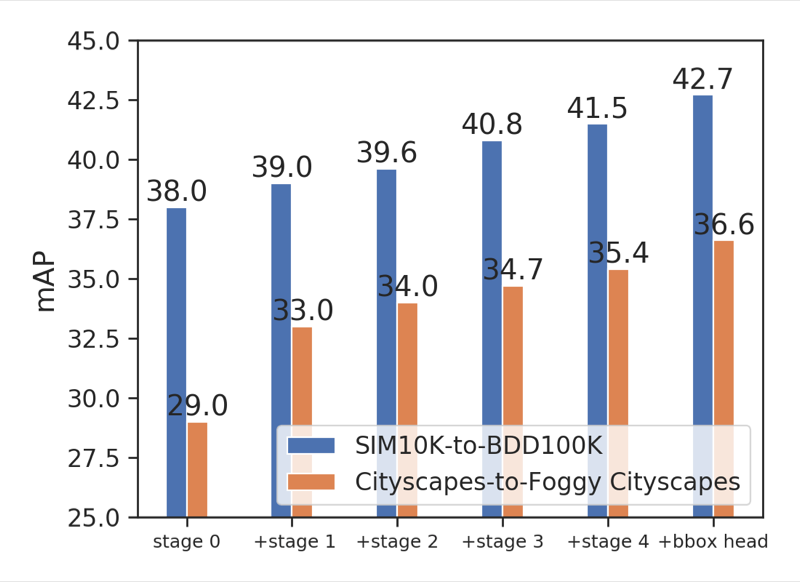}}
    \subfloat[]{\includegraphics[width=0.345\linewidth]{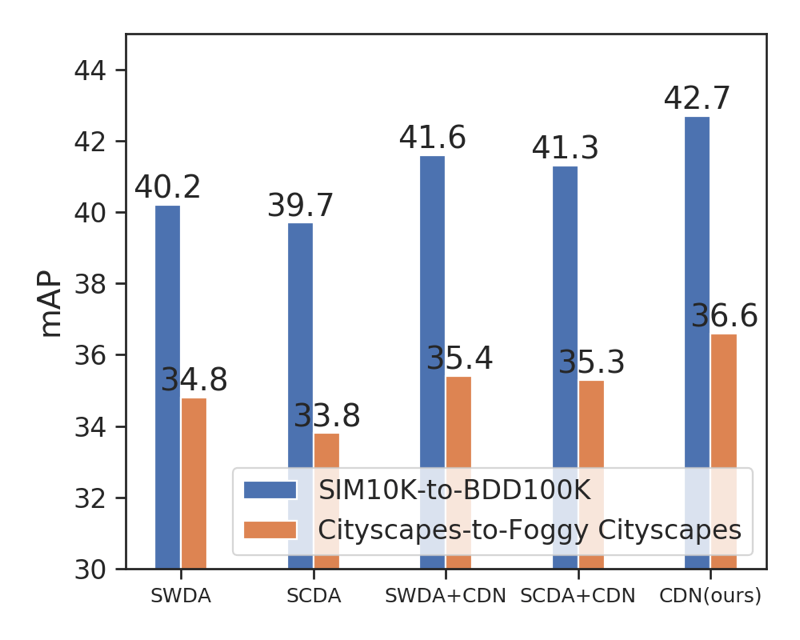}}
    \subfloat[]{\includegraphics[width=0.275\linewidth]{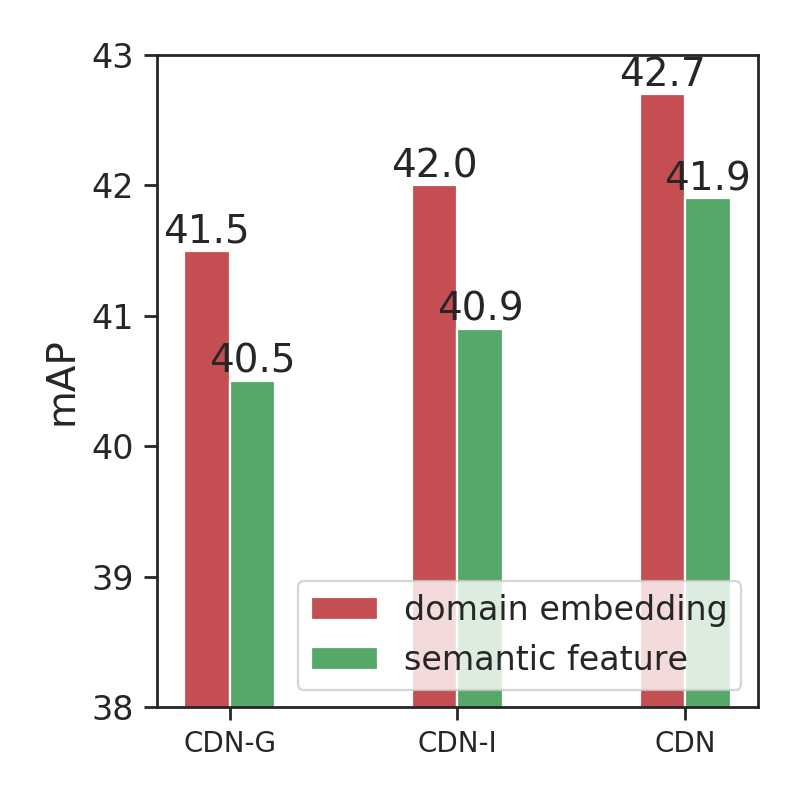}}
    \\ [-1.3ex]
  \caption{(a) Adopt CDN at different convolution stages of ResNet; (b) Adopt CDN in existing adaptation frameworks; (c) Domain embedding vs. semantic features.}
  \label{fig:3-ablations}
\end{figure}

\textbf{Adopting CDN at different convolution stages}.
Fig.~\ref{fig:3-ablations}(a) compares the results of Faster R-CNN models adopting CDN at different convolution stages.
We follow \cite{he2016deep} to divide ResNet into $5$ stages.
Bbox head denotes the bounding box head network.
From left to right, adding more CDN layers keeps boosting the adaptation performance on both benchmarks, benefiting from adaptive distribution alignments across different levels' representation.
It suggests that adopting CDN in each convolution stage is a better choice than only aligning domain distributions at one or two specific convolution stages.

\textbf{Comparing with existing domain adaptation frameworks adopting CDN}.
Fig.~\ref{fig:3-ablations}(b) shows the results of adopting CDN layer in existing adaptation methods like SWDA~\cite{Saito_2019_CVPR} and  SCDA~\cite{Zhu_2019_CVPR}.
Directly adopting CDN in SWDA and SCDA can bring average $1.3\%$ mAP improvements on two adaptation benchmarks, suggesting CDN is more effective to address domain shifts than traditional domain confusion learning.
It can be attributed to that CDN disentangle the domain-specific factors out of the semantic features via learning a domain-vector.
Leveraging the domain-vector to align the different domain distributions can be more efficient.

\textbf{Compare domain embedding with semantic features}.
In Eq.~\ref{eq:CDN_loss}, we can either use  semantic features $(v^s, \hat v^t)$ or  domain embedding $(\mathbf{F}_d(v^s), \mathbf{F}_d(\hat v^t))$ as inputs of discriminator.
Fig.~\ref{fig:3-ablations}(c) compares the adaptation performance of using semantic features with using domain embedding.
Although semantic features can improve the performance over baseline, domain embedding consistently achieves better results than directly using semantic features.
Suggesting the learned domain embedding well captures the domain attribute information, and it is free from some undesirable regularization on specific image contents.

\begin{figure}
    \centering
    \includegraphics[width=0.9\linewidth]{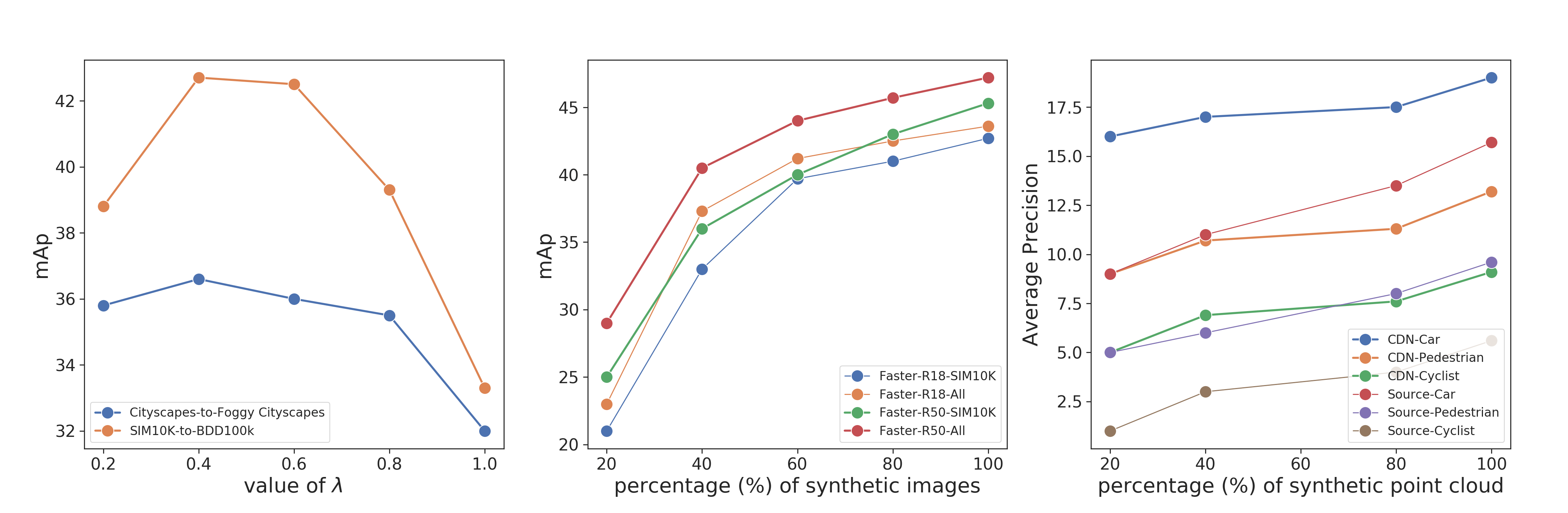}
    \caption{Left: mAP vs. Value of $\lambda$; Middle:  mAP vs. Percentage (\%) of synthetic image data;  Right:  AP vs. Percentage (\%) of synthetic point cloud.}
    \label{fig:ablation-lamda-cg}
\end{figure}

\textbf{Value of $\lambda$}
In Eq.~\ref{eq:overall_objective}, we use $\lambda$ controls the balance between global and local regularization.
Fig.~\ref{fig:ablation-lamda-cg} (left) shows the influence on adaptation performance by different $\lambda$.
Because object detectors naturally focus more on local regions, we can see stronger instance regularization largely contributes to detection performance.
In our experiments, $\lambda$ between 0.4 and 0.5 gives the best performance.

\textbf{Scale of target domain dataset}
Fig.~\ref{fig:ablation-lamda-cg} middle/right quantitatively investigate the relation between real data detection performance and percentage of synthetic data used for training.
``All''  means to use the combination of 3 different synthetic datasets. The larger synthetic dataset provides better adaptation performance, on both 2D image and 3D point cloud detection.

\section{Conclusion}
We present the Conditional Domain Normalization (CDN) to adapt object detectors across different domains.
CDN aims to embed different domain inputs into a shared latent space, where the features from different domains carry the same domain attribute.
Extensive experiments demonstrate the effectiveness of CDN on adapting object detectors,  including  2D image and 3D point cloud detection tasks.
And both quantitative and qualitative comparisons are conducted to analyze the features learned by our method.

\vspace{30pt}
\noindent \textbf{\Large Appendix}
\vspace{8pt}

\noindent  \textbf{A1. Interpret the  Domain Embedding}

\begin{figure*}[]
\centering
\subfloat{\raisebox{0.35in}{\rotatebox[origin=t]{90}{\tiny{Original} }}} \hspace{0.1pt}  
\subfloat{\includegraphics[width=1.7\widthdef]{sup_fig/foggy2city/epoch160_real_B.png}}  \hspace{0.3pt}
\subfloat{\includegraphics[width=1.7\widthdef]{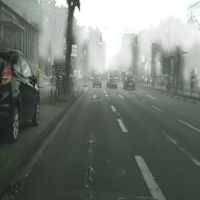}}     \hspace{0.3pt}
\subfloat{\includegraphics[width=1.7\widthdef]{sup_fig/foggy2city/epoch170_real_B.png}}   \hspace{0.3pt}
\subfloat{\includegraphics[width=1.7\widthdef]{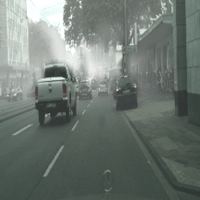}}   \hspace{0.3pt}
\subfloat{\includegraphics[width=1.7\widthdef]{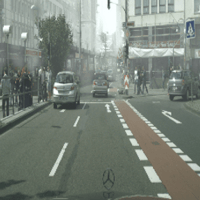}}    \\

\subfloat{\raisebox{0.39in}{\rotatebox[origin=t]{90}{{\tiny{ Original's feature} }}}} \hspace{0.1pt}
\subfloat{\includegraphics[width=1.7\widthdef]{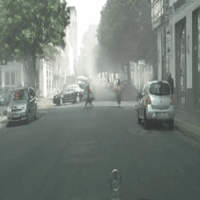}}  \hspace{0.3pt}
\subfloat{\includegraphics[width=1.7\widthdef]{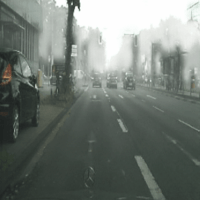}}     \hspace{0.3pt}
\subfloat{\includegraphics[width=1.7\widthdef]{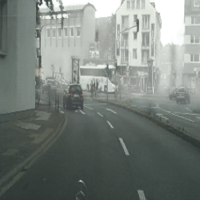}}   \hspace{0.3pt}
\subfloat{\includegraphics[width=1.7\widthdef]{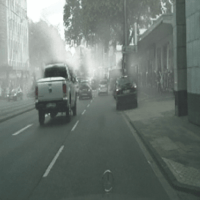}}   \hspace{0.3pt}
\subfloat{\includegraphics[width=1.7\widthdef]{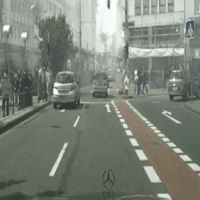}}   \\ \vspace{-5pt}

\subfloat{\raisebox{0.36in}{\rotatebox[origin=t]{90}{ \tiny{+Domain Embedding} }}} \hspace{0.1pt}
\subfloat{\includegraphics[width=1.7\widthdef]{sup_fig/foggy2city/epoch160_fake_A.png}}  \hspace{0.3pt}
\subfloat{\includegraphics[width=1.7\widthdef]{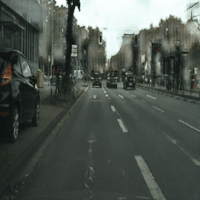}}     \hspace{0.3pt}
\subfloat{\includegraphics[width=1.7\widthdef]{sup_fig/foggy2city/epoch170_fake_A.png}}   \hspace{0.3pt}
\subfloat{\includegraphics[width=1.7\widthdef]{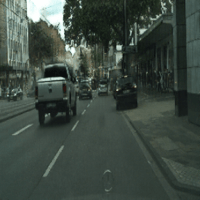}}   \hspace{0.3pt}
\subfloat{\includegraphics[width=1.7\widthdef]{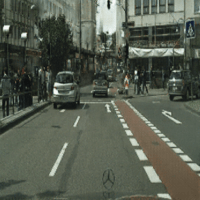}}     \\ \vspace{-3pt}

\caption{Top row: Original inputs of Foggy Cityscapes;
Middle row: Reconstructed results from features of original inputs;
Bottom row: Reconstructed results from features encoded with the domain embedding of Cityscapes.
}
\label{fig:fogggy2city_domain_embedding}
\end{figure*}

\begin{figure*}[]
\centering
\subfloat{\raisebox{0.35in}{\rotatebox[origin=t]{90}{\tiny{Original} }}} \hspace{0.1pt}
\subfloat{\includegraphics[width=1.7\widthdef]{sup_fig/gta2bdd/epoch138_real_A.png}}  \hspace{0.3pt}
\subfloat{\includegraphics[width=1.7\widthdef]{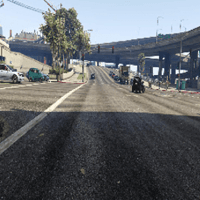}}     \hspace{0.3pt}
\subfloat{\includegraphics[width=1.7\widthdef]{sup_fig/gta2bdd/epoch156_real_A.png}}   \hspace{0.3pt}
\subfloat{\includegraphics[width=1.7\widthdef]{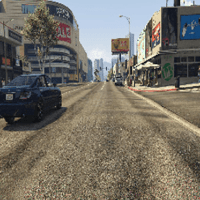}}   \hspace{0.3pt}
\subfloat{\includegraphics[width=1.7\widthdef]{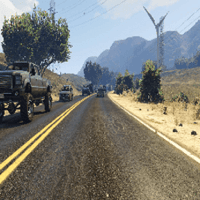}}     \\

\subfloat{\raisebox{0.38in}{\rotatebox[origin=t]{90}{\tiny{Original's feature} }}} \hspace{0.1pt}
\subfloat{\includegraphics[width=1.7\widthdef]{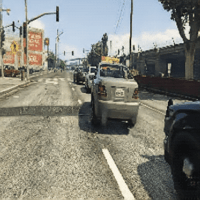}}  \hspace{0.3pt}
\subfloat{\includegraphics[width=1.7\widthdef]{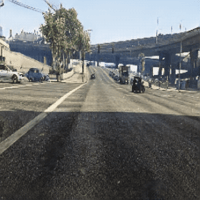}}     \hspace{0.3pt}
\subfloat{\includegraphics[width=1.7\widthdef]{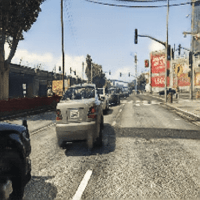}}   \hspace{0.3pt}
\subfloat{\includegraphics[width=1.7\widthdef]{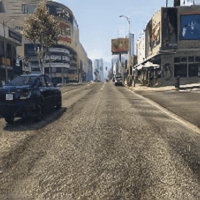}}   \hspace{0.3pt}
\subfloat{\includegraphics[width=1.7\widthdef]{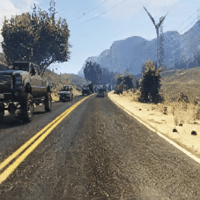}}     \\ \vspace{-3pt}

\subfloat{\raisebox{0.36in}{\rotatebox[origin=t]{90}{\tiny{+Domain Embedding} }}} \hspace{0.1pt}
\subfloat{\includegraphics[width=1.7\widthdef]{sup_fig/gta2bdd/epoch138_fake_B.png}}  \hspace{0.3pt}
\subfloat{\includegraphics[width=1.7\widthdef]{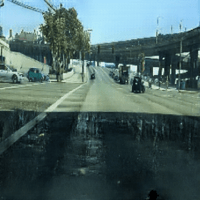}}     \hspace{0.3pt}
\subfloat{\includegraphics[width=1.7\widthdef]{sup_fig/gta2bdd/epoch156_fake_B.png}}   \hspace{0.3pt}
\subfloat{\includegraphics[width=1.7\widthdef]{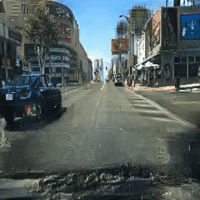}}   \hspace{0.3pt}
\subfloat{\includegraphics[width=1.7\widthdef]{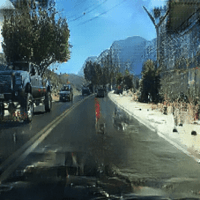}}     \\

\caption{Top row: Original inputs of SIM10K;
Middle row: Reconstructed results from features of original inputs;
Bottom row: Reconstructed results from features encoded with the domain embedding of BDD100K.
}
\label{fig:gta2bdd_domain_embedding}
\end{figure*}

\begin{figure}
    \centering
    \includegraphics[width=0.9\linewidth]{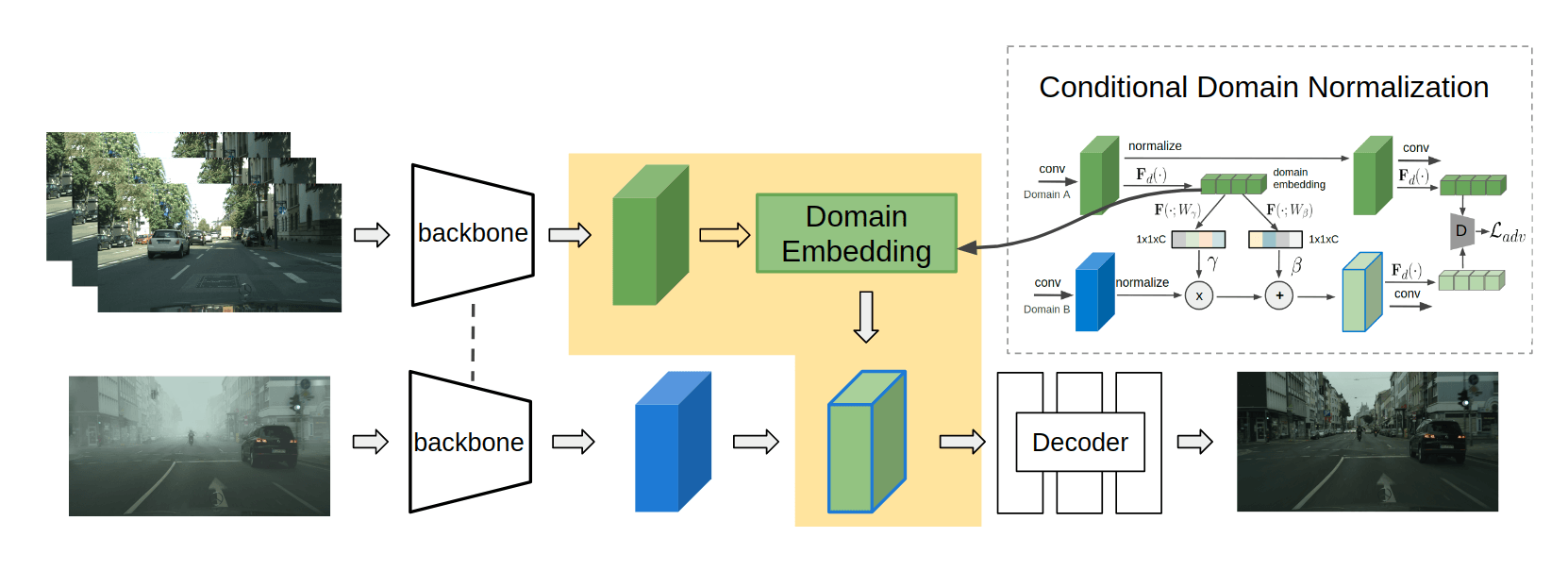}
    \caption{Interpreting the learned domain embedding with a decoder network.}
    \label{fig:cdn-decoder}
\end{figure}

Conditional domain normalization disentangles the domain-specific attribute out of the semantic features from one domain via a learning a domain embedding to characterize the domain attribute information.
In this section, we interpret the learned domain embedding via reconstructing the RGB images from the features.
As shown in Fig.~\ref{fig:cdn-decoder}, we first built a decoder network $Decoder(\cdot; \theta_{dec})$ upon the backbone network $G(\cdot; \theta_{g}^*)$ of fixed weights.
The parameters of the backbone network are obtained in the adaptation training (see Eq.~\ref{eq:1}).
The decoder network mostly mirrors the backbone network, with all pooling layers replaced by nearest up-sampling and all normalization layers removed.
The decoder network is trained to reconstruct the RGB images from the features extracted by the backbone,
\begin{equation}
   \arg \min_{ \theta_{dec}}  \mathcal{L} = || Decoder( G(x;  \theta_{g}^*);  \theta_{dec}) - x ||_2.
\end{equation}
For contrast analysis, only single domain images are used to train the decoder network, i.e. the decoder for Cityscapes experiment is trained on Foggy Cityscapes images, the decoder for SIM10K experiment is trained on SIM10K images.
After we got a trained decoder network, we use it to reconstruct the RGB image from features encoded with the domain embedding.

Fig.~\ref{fig:fogggy2city_domain_embedding} shows the effect of domain embedding learned in Cityscapes to Foggy Cityscapes adaptation experiments (Section~\ref{sec:5-1}).
The top row shows the inputs of Foggy Cityscapes;
the middle row shows the reconstructed results from features of Foggy Cityscapes inputs;
the bottom row is reconstructed results from Foggy Cityscapes features encoded with the domain embedding learned on Cityscapes.
With the help of the domain embedding learned on Cityscapes, the reconstructed results from Foggy Cityscapes features no longer exhibit foggy characteristics, suggesting that both Cityscapes and Foggy Cityscapes inputs are embedded into a shared latent space, where their features carry the same domain attribute.
Given the domain gap bridged, the object detector supervised trained on Cityscapes also works on Foggy Cityscapes.

\begin{figure*}[]
\centering
\subfloat{\raisebox{0.35in}{\rotatebox[origin=t]{90}{\tiny{Original} }}} \hspace{0.1pt}
\subfloat{\includegraphics[width=1.7\widthdef]{sup_fig/syn2bdd/epoch101_real_A.png}}  \hspace{0.3pt}
\subfloat{\includegraphics[width=1.7\widthdef]{sup_fig/syn2bdd/epoch112_real_A.png}}     \hspace{0.3pt}
\subfloat{\includegraphics[width=1.7\widthdef]{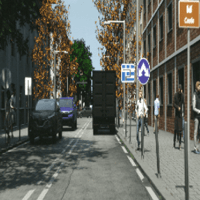}}   \hspace{0.3pt}
\subfloat{\includegraphics[width=1.7\widthdef]{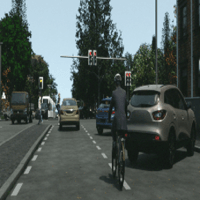}}   \hspace{0.3pt}
\subfloat{\includegraphics[width=1.7\widthdef]{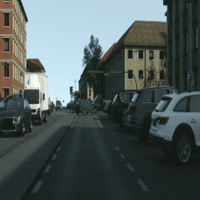}}     \\

\subfloat{\raisebox{0.38in}{\rotatebox[origin=t]{90}{\tiny{Original's feature} }}} \hspace{0.1pt}
\subfloat{\includegraphics[width=1.7\widthdef]{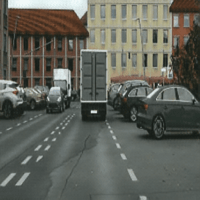}}  \hspace{0.3pt}
\subfloat{\includegraphics[width=1.7\widthdef]{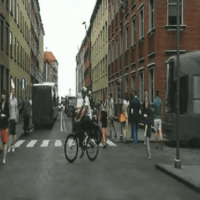}}     \hspace{0.3pt}
\subfloat{\includegraphics[width=1.7\widthdef]{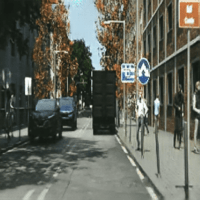}}   \hspace{0.3pt}
\subfloat{\includegraphics[width=1.7\widthdef]{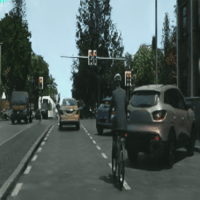}}   \hspace{0.3pt}
\subfloat{\includegraphics[width=1.7\widthdef]{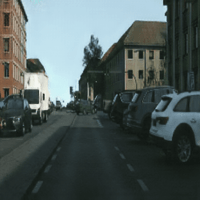}}     \\ \vspace{-3pt}

\subfloat{\raisebox{0.36in}{\rotatebox[origin=t]{90}{\tiny{+Domain Embedding} }}} \hspace{0.1pt}
\subfloat{\includegraphics[width=1.7\widthdef]{sup_fig/syn2bdd/epoch101_fake_B.png}}  \hspace{0.3pt}
\subfloat{\includegraphics[width=1.7\widthdef]{sup_fig/syn2bdd/epoch112_fake_B.png}}     \hspace{0.3pt}
\subfloat{\includegraphics[width=1.7\widthdef]{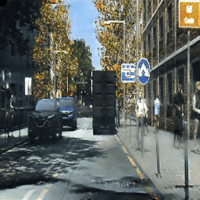}}   \hspace{0.3pt}
\subfloat{\includegraphics[width=1.7\widthdef]{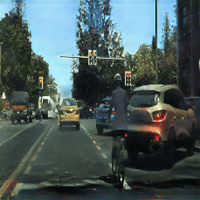}}   \hspace{0.3pt}
\subfloat{\includegraphics[width=1.7\widthdef]{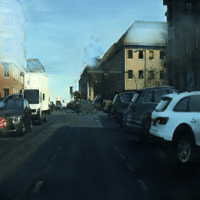}}     \\

\caption{Top row: Original inputs of Synscapes;
Middle row: Reconstructed results from features of original inputs;
Bottom row: Reconstructed results from features encoded with the domain embedding of BDD100K.
}
\label{fig:synscapes2bdd_domain_embedding}
\end{figure*}

Fig.~\ref{fig:gta2bdd_domain_embedding} and~\ref{fig:synscapes2bdd_domain_embedding} show the reconstructed results from synthetic data's features encoded with domain embedding of real data (BDD100K), which are learned in SIM10K-to-BDD100K and Synscapes-to-BDD10K adaptation experiments, respectively (see Section~\ref{sec:5-4}).
Without the domain embedding of real data, the reconstructed images (middle row of Fig~\ref{fig:gta2bdd_domain_embedding} and~\ref{fig:synscapes2bdd_domain_embedding}) still exhibit characteristic of CG (computer graphic), that look identical to the original images.
When the same features of original inputs are encoded with the domain embedding of real data, the reconstructed images (bottom row of Fig~\ref{fig:gta2bdd_domain_embedding} and~\ref{fig:synscapes2bdd_domain_embedding}) obviously becomes more realistic.
For example, the color of the sky, the texture of the road and objects in the reconstructed images look similar to the real images.
It proves that the learned domain embedding well captures the domain attribute information of real data, and it can be used to effectively translate the synthetic images towards real images.

\vspace{15pt}

\begin{figure}
    \centering
    \includegraphics[width=0.93\linewidth]{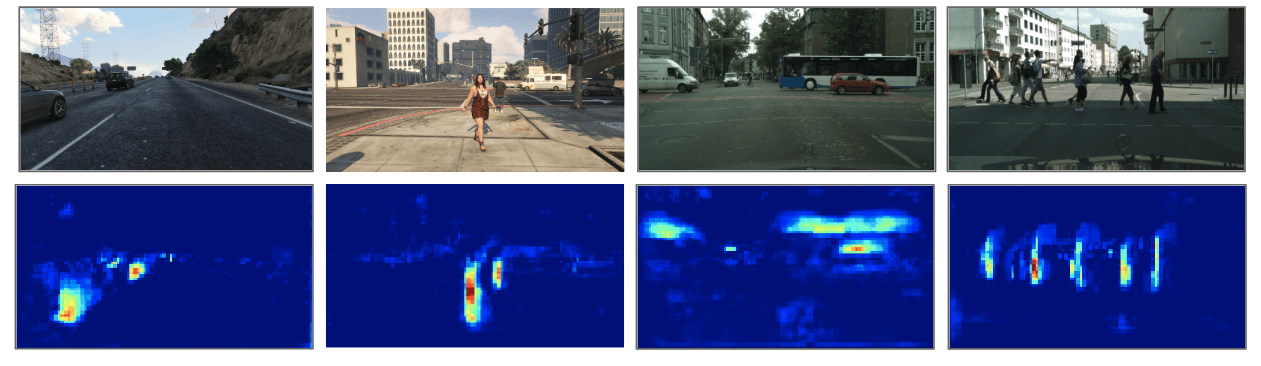}
    \caption{Res5-3 features learned by Faster R-CNN with CDN. Left two images are from synthetic data (SIM10K)  and the right two images are from real data (Cityscapes).}
    \label{fig:CDN-feature}
\end{figure}

\noindent \textbf{A.2 Visualize the Feature Maps}
Despite the general efficiency on various benchmarks, we are also interested in the underlying principle of CDN.
We first visualize the features of different domain images.
As shown in Fig.~\ref{fig:CDN-feature}, we can not easily distinguish the domain label from feature maps alone, suggesting the features from synthetic and real domain carry the same domain attribute.
Besides, the same category objects across synthetic and real domain share similar activation patterns and contours, indicating the our method well preserves the feature semantics.

\vspace{15pt}

\begin{figure*}[]
\centering
\subfloat{\includegraphics[width=2.3\widthdef]{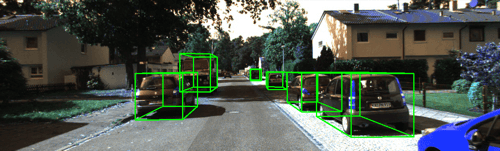}}  \hspace{0.3pt}
\subfloat{\includegraphics[width=2.3\widthdef]{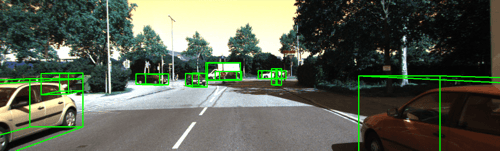}}     \hspace{0.3pt}
\subfloat{\includegraphics[width=2.3\widthdef]{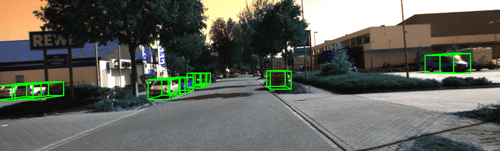}}   \hspace{0.3pt}
\subfloat{\includegraphics[width=2.3\widthdef]{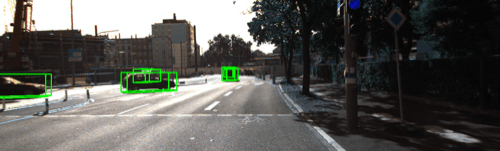}}     \\

\subfloat{\includegraphics[width=2.3\widthdef]{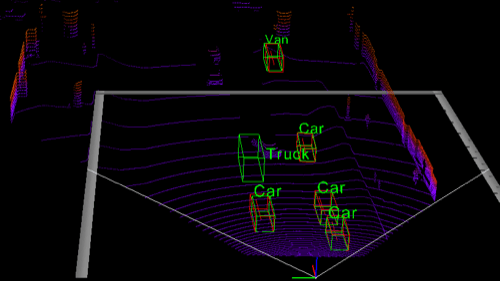}}  \hspace{0.3pt}
\subfloat{\includegraphics[width=2.3\widthdef]{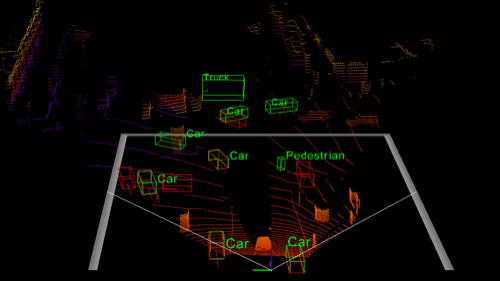}}     \hspace{0.3pt}
\subfloat{\includegraphics[width=2.3\widthdef]{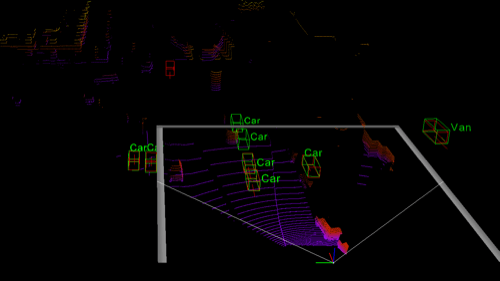}}   \hspace{0.3pt}
\subfloat{\includegraphics[width=2.3\widthdef]{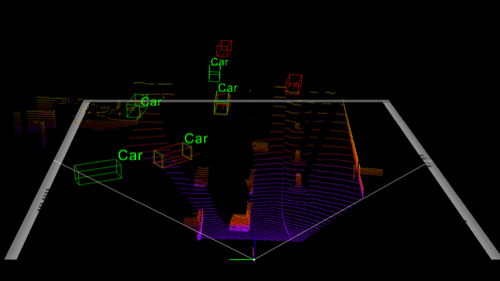}}     \\

\subfloat{\includegraphics[width=2.3\widthdef]{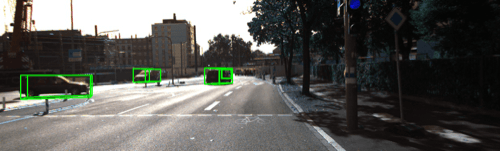}}  \hspace{0.3pt}
\subfloat{\includegraphics[width=2.3\widthdef]{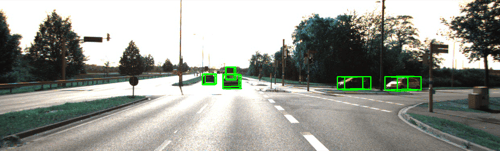}}     \hspace{0.3pt}
\subfloat{\includegraphics[width=2.3\widthdef]{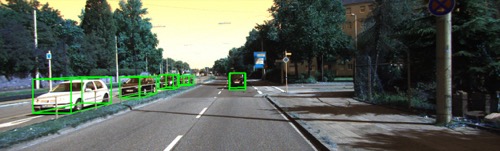}}   \hspace{0.3pt}
\subfloat{\includegraphics[width=2.3\widthdef]{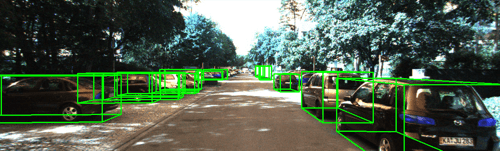}}     \\

\subfloat{\includegraphics[width=2.3\widthdef]{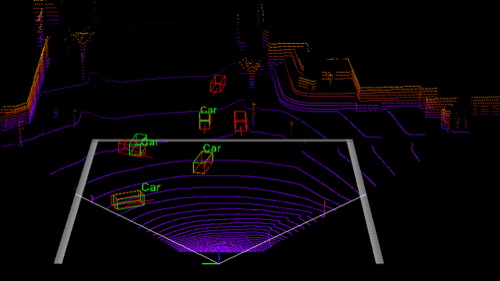}}  \hspace{0.3pt}
\subfloat{\includegraphics[width=2.3\widthdef]{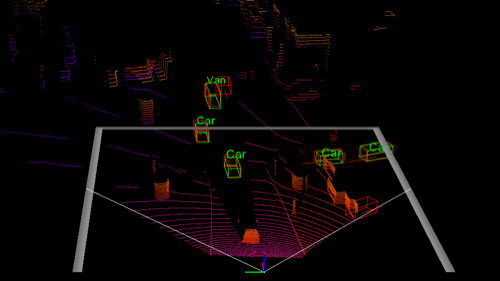}}     \hspace{0.3pt}
\subfloat{\includegraphics[width=2.3\widthdef]{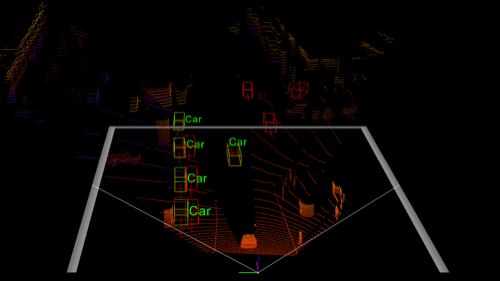}}   \hspace{0.3pt}
\subfloat{\includegraphics[width=2.3\widthdef]{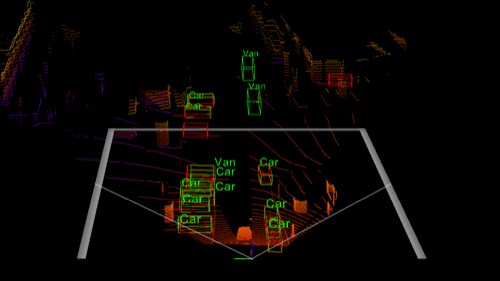}}     \\

\caption{Example results of PointRCNN model with Conditional Domain Normalization (with AP of moderate level of $19.0$). For each example, the upper part is the image and the lower part is the corresponding point cloud. The detected objects are shown with red 3D bounding boxes. The green bounding boxes represent the ground truth.
}
\label{fig:3d-kitti}
\end{figure*}

\noindent \textbf{A.3 More Qualitative Results}
Fig.~\ref{fig:3d-kitti} shows 3D point cloud detection results on the KITTI dataset. \\

\bibliographystyle{splncs04}
\bibliography{refs}

\end{document}